\documentclass[sigconf,nonacm]{acmart}

\AtBeginDocument{%
  }

\copyrightyear{}
\acmYear{}
\acmDOI{}

\acmConference[]{}{}{}
\acmBooktitle{}
\acmISBN{}

\usepackage{multirow}

\newcommand{\yq}[1]{
    \textcolor{black}{{#1}}
}

\newcommand{\ourmethod}{CT-NeRF}
\newcommand{\ie}{\emph{i.e.} }
\begin{document}

\newcommand{\reflabel}{dummy} 

\newcommand{\be}{\begin{equation}}
\newcommand{\ee}{\end{equation}}
\newcommand{\eqlabel}[1]{\label{eq:\reflabel-#1}}
\renewcommand{\eqref}[2][\reflabel]{Eq. (\ref{eq:#1-#2})}

\newcommand{\seclabel}[1]{\label{sec:\reflabel-#1}}
\newcommand{\secref}[2][\reflabel]{Section~\ref{sec:#1-#2}}
\newcommand{\Secref}[2][\reflabel]{Section~\ref{sec:#1-#2}}
\newcommand{\secrefs}[3][\reflabel]{Sections~\ref{sec:#1-#2} and~\ref{sec:#1-#3}}

\newcommand{\Eqref}[2][\reflabel]{(\ref{eq:#1-#2})}
\newcommand{\eqrefs}[3][\reflabel]{(\ref{eq:#1-#2}) and~(\ref{eq:#1-#3})}

\newcommand{\figlabel}[2][\reflabel]{\label{fig:#1-#2}}
\newcommand{\figref}[2][\reflabel]{Fig.~\ref{fig:#1-#2}}
\newcommand{\Figref}[2][\reflabel]{Fig.~\ref{fig:#1-#2}}
\newcommand{\figsref}[3][\reflabel]{Figs.~\ref{fig:#1-#2} and~\ref{fig:#1-#3}}
\newcommand{\Figsref}[3][\reflabel]{Figs.~\ref{fig:#1-#2} and~\ref{fig:#1-#3}}

\newcommand{\alglabel}[2][\reflabel]{\label{alg:#1-#2}}
\newcommand{\algref}[2][\reflabel]{Algorithm~\ref{alg:#1-#2}}
\newcommand{\Algref}[2][\reflabel]{Algorithm~\ref{alg:#1-#2}}

\newcommand{\tablelabel}[2][\reflabel]{\label{table:#1-#2}}
\newcommand{\tableref}[2][\reflabel]{Table~\ref{table:#1-#2}}
\newcommand{\Tableref}[2][\reflabel]{Table~\ref{table:#1-#2}}

\def\bfmu{\mbox{\boldmath$\mu$}}
\def\bftau{\mbox{\boldmath$\tau$}}
\def\bftheta{\mbox{\boldmath$\theta$}}
\def\bfdelta{\mbox{\boldmath$\delta$}}
\def\bfphi{\mbox{\boldmath$\phi$}}
\def\bfpsi{\mbox{\boldmath$\psi$}}
\def\bfeta{\mbox{\boldmath$\eta$}}
\def\bfnabla{\mbox{\boldmath$\nabla$}}
\def\bfGamma{\mbox{\boldmath$\Gamma$}}

%



\newcommand{\R}{\mathbb{R}}

\title{CT-NeRF: Incremental Optimizing Neural Radiance Field and Poses with Complex Trajectory}

\author{Yunlong Ran}
\authornote{Both authors contributed equally to the paper. Email: yunlong\_ran@zju.edu.cn}
\affiliation{%
  \institution{Zhejiang University}
  \city{Hangzhou}
  \country{China}
}

\author{Yanxu Li}
\authornotemark[1]
\affiliation{%
  \institution{Zhejiang University}
  \city{Hangzhou}
  \country{China}
}

\author{Qi Ye}
\authornote{Corresponding author. Email: qi.ye@zju.edu.cn}
\affiliation{
  \institution{Zhejiang University}
  \city{Hangzhou}
  \country{China}
}

\author{Yuchi Huo}
\affiliation{
  \institution{Zhejiang University}
  \city{Hangzhou}
  \country{China}
}

\author{Zechun Bai}
\affiliation{%
  \institution{Shandong University}
  \city{Jinan}
  \country{China}
}

\author{Jiahao sun}
\affiliation{%
  \institution{Zhejiang University}
  \city{Hangzhou}
  \country{China}
}

\author{Jiming chen}
\affiliation{
  \institution{Zhejiang University}
  \city{Hangzhou}
  \country{China}
}

\renewcommand{\shortauthors}{}

\begin{abstract}
Neural radiance field (NeRF) has achieved impressive results in high-quality 3D scene reconstruction. However, NeRF heavily relies on precise camera poses. While recent works like BARF have introduced camera pose optimization within NeRF, their applicability is limited to simple trajectory scenes. Existing methods struggle while tackling complex trajectories involving large rotations. To address this limitation, we propose ~\ourmethod{}, an incremental reconstruction optimization pipeline using only RGB images without pose and depth input. In this pipeline, we first propose a local-global bundle adjustment under a pose graph connecting neighboring frames to enforce the consistency between poses to escape the local minima caused by only pose consistency with the scene structure. Further, we instantiate the consistency between poses as a reprojected geometric image distance constraint resulting from pixel-level correspondences between input image pairs. Through the incremental reconstruction, ~\ourmethod{} enables the recovery of both camera poses and scene structure and is capable of handling scenes with complex trajectories. We evaluate the performance of ~\ourmethod{} on two real-world datasets, NeRFBuster and Free-Dataset, which feature complex trajectories. Results show ~\ourmethod{} outperforms existing methods in novel view synthesis and pose estimation accuracy.
\end{abstract}



\begin{teaserfigure}
  \centering
  \includegraphics[width=0.9\textwidth]{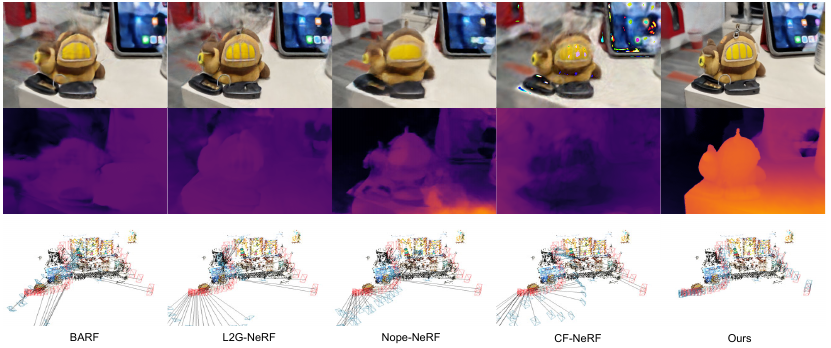}
  \caption{Comparison on Free-dataset~\cite{wang2023f2}. 
  Top: novel view synthesis; Middle: depth maps; Bottom: the estimated trajectory errors (red rectangles for poses from COLMAP, blue rectangles for estimated, gray lines between them for errors). Our method enables more robust pose estimation, renders better novel views, and constructs better geometry than the state-of-the-arts.}
  \Description{Enjoying the baseball game from the third-base
  seats. Ichiro Suzuki preparing to bat.}
  \figlabel{teaser}
\end{teaserfigure}

\maketitle

\section{Introduction}
\label{sec:intro}

\begin{figure*}[!htbp]
     \centering
     \includegraphics[width=0.95\textwidth]{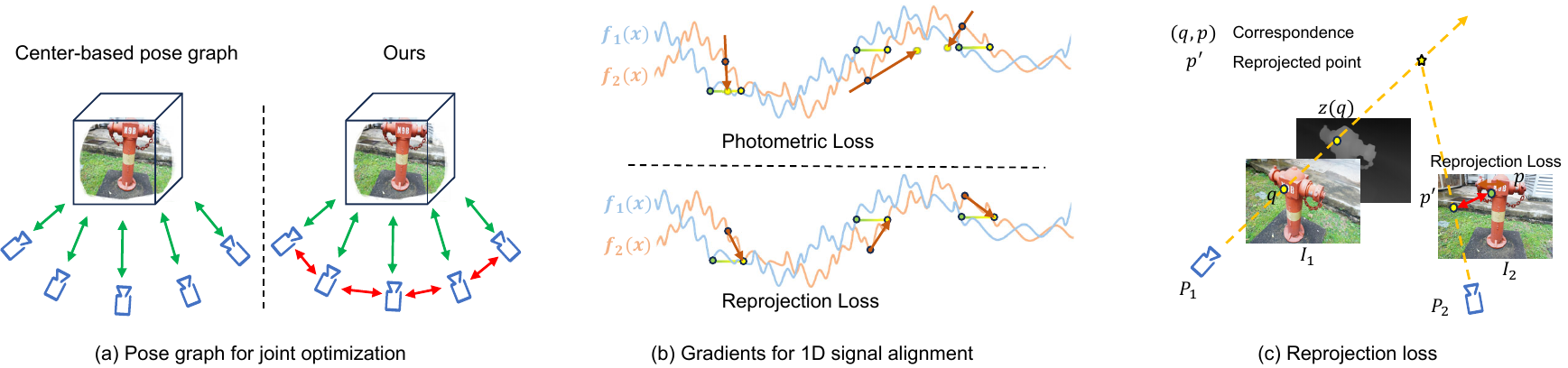}
     \caption{(a) Left: center-based pose graph to force pose consistent to the scene; Right: our pose graph to enable consistency between the camera poses in addition to the consistency to the scene. (b) The reprojection loss (bottom) provides an accurate gradient towards alignment, while the photometric loss (top) provides inconsistent gradients. (c) For a pair of correspondence $(q,p)$ between $I_1$ and $I_2$, $q$ is reprojected to the image plane of $I_2$  via the depth value of $q$. The reprojection loss is the geometric image distance between the reprojected point $p'$ and the ground truth corresponding point $p$.}
\figlabel{combine}
\end{figure*}

Reconstructing high-fidelity, high-quality 3D scenes holds significant importance for the development of virtual reality / augmented reality, autonomous driving, and other domains. Recently, implicit representations such as neural radiance fields (NeRF)~\cite{mildenhall2021nerf} have achieved remarkable progress in reconstructing photo-realistic scenes given a sequence of RGB images and their corresponding camera poses. The camera poses for these high-fidelity reconstructions with implicit representations are primarily obtained through the structure from motion (SfM) methods, with the off-the-shelf tool COLMAP~\cite{schonberger2016structure} being the most popular choice. These SfM methods estimate the camera poses through local registration between images and global bundle adjustment (BA) on both all camera poses and sparse 3D scene points. Therefore, accurate camera poses can only be acquired after all images are processed. Also, matching and registration of the methods are sensitive to image variations.

Recent works such as NeRFmm~\cite{wang2021nerf}, SC-NeRF~\cite{jeong2021self}, BARF~\cite{lin2021barf}, GARF~\cite{chng2022gaussian} and L2G-NeRF~\cite{chen2023local} tackle the dependency on the camera pose priors by treating the camera pose as learnable parameters and jointly optimizing poses and scenes offline. However, using images only to constrain the 3D space encounters many problems like wrong geometry, blurry textures, or floaters when very dense multiview images are not available; adding more freedoms for the camera poses to the optimization leads to worse results. Therefore, these methods often require initial camera parameters close to the ground truth poses in object-centered scenes with dense multiview observations, or small camera movements.
Nope-NeRF~\cite{bian2023nope} incorporates monocular depth to impose further constraints on adjacent images, enabling pose estimation for trajectories with relatively small camera motions and rotations while its initialization of all poses as identity matrices leads to local optima when facing complex trajectories.

On the other hand, following the classic SfM pipelines, CF-NeRF~\cite{yan2023cfnerf}  adds images incrementally, initializes the pose for a newly added image with the pose for the previous one, and optimizes the poses (and the scene). With the incremental strategy, the method is capable of reconstructing the real-world scene under complex camera trajectories. However, it still suffers from large pose errors and inferior reconstruction quality as shown in~\figref{teaser}, ~\figref{rgbcmpfree} and ~\figref{cmptraj}. The reason can be attributed to two aspects. 
First, the bundle adjustment constructs a center-based graph as shown in~\figref{combine} (a) left,  optimizing only the consistency between the camera poses and the center implicit global scene while neglecting the consistency between the pose and the multiview images. When the global structure falls into local minima, the camera poses cannot be recovered; in turn, the structure cannot find a way to escape the local minima as the poses and structure are optimized jointly.
Secondly, the method only uses the visual difference between the rendering images and raw images whereas BARF~\cite{lin2021barf} observes that as natural images are typically complex signals, gradient-based registration with pixel value differences is susceptible to suboptimal solutions if poorly initialized, as shown in the top figure of ~\figref{combine} (b). The coarse-to-fine pose estimation proposed by BARF~\cite{lin2021barf} mitigates this issue but requires good initialization or dense forward-facing images. In complex trajectories, the camera typically exhibits large motions, and views covering a region are much sparser than the forward-facing scenario. 

To tackle the issues and enable accurate pose estimation and reconstruction for complex trajectories, we propose a novel incremental joint optimization method for implicit radiance fields and camera poses named~\ourmethod. 
For the first issue, as shown in~\figref{combine} (a) right, we propose a joint incremental reconstruction and pose estimation pipeline with pose graphs connecting edges between camera poses upon the center-based pose graphs for a local-global bundle adjustment (BA). The graph forms many subgraphs and forces consistency between the camera poses, which helps to recover the poses when the scene and poses are consistent but the scene is actual in a local minimum during BA. 
For the second issue and also instantiating the pose consistency between the pose edges, we introduce a geometric image instance, \ie the reprojected Euclidean distance between the correspondences of two input images. In addition to providing consistency constraints for pose edges, the reprojected distance benefits the pose and scene optimization in three aspects: 1) it provides direct direction to align the poses, whereas the pixel value differences do not necessarily correlate to the pose error: as shown in~\figref{combine} (b) gradients based on the pixel value difference are not consistent while the gradients from the reprojected geometric image distance are; 2) the reprojected distance requires the depth of the scene to warp a pixel in an image to the other one as shown in~\figref{combine} (c) and therefore, the gradient can help the convergence of the geometry of the scene directly; 3)  correspondence learning networks typically leverage large scale pair-wise image datasets and the reprojected loss based on the correspondences is robust to occlusion, lighting variation, textureless and large motions compared to raw image losses.

In summary, our main contributions are threefold: 
\begin{itemize}
    \item We design an incremental reconstruction pipeline for neural radiance fields using only RGB images under complex camera trajectories, without pose and depth input.  
    \item We propose to construct pose graphs with in-between pose consistency edges for BA and instantiate the consistency as a reprojected geometric image distance constraint from the learned correspondences between input images for robust pose and scene optimization.
    \item We achieve significant improvements in pose estimation accuracy and reconstruction quality compared to state-of-the-art methods in complex trajectories.
\end{itemize}

\section{Related Work}

\noindent\textbf{SFM and SLAM} In the field of computer vision, given a set of input images, SfM and SLAM aim to concurrently estimate camera poses and reconstruct the scene. The distinction lies in the fact that SLAM operates online, emphasizing runtime performance, while SfM does not require online operation but demands higher accuracy. SfM methods can be categorized into incremental~\cite{schonberger2016structure,wu2013towards,snavely2008modeling}, global~\cite{jiang2013global,cui2015global}, and hierarchical~\cite{gherardi2010improving} approaches: The incremental approach initializes with two images and progressively registers and reconstructs additional images one by one. The global approach registers and reconstructs all images simultaneously. The hierarchical approach first groups images, performs registration and reconstruction for each group, and then conducts a global optimization. SLAM methods are primarily divided into filter-based~\cite{bailey2006consistency,castellanos2007robocentric,abdelrasoul2016quantitative} and graph optimization-based~\cite{engel2014lsd,mur2015orb,campos2021orb} approaches. Filter-based methods mainly utilize state estimation strategies such as Kalman filtering and particle filtering to incrementally estimate the posterior distributions of camera poses and key point locations. Graph optimization-based methods abstract camera poses at different times as nodes and the observation constraints at different robot locations as edges connecting the nodes, then employ bundle adjustment (BA) algorithms for global optimization.
Our proposed incremental pipeline is inspired by the incremental SfM and SLAM approaches.

\noindent\textbf {NeRF-based SFM and SLAM} Implicit neural representations have gained prominence since 2019~\cite{park2019deepsdf}. Compared to traditional explicit representations that store geometric information in a relatively fixed and simple manner, implicit neural representations can better handle complex topological structures and geometric details. The classic algorithm for implicit neural representations, Vanilla NeRF~\cite{mildenhall2021nerf} (Neural Radiance Fields), is based on the theory of volume rendering and utilizes a Multi-Layer Perceptron (MLP) to learn the implicit neural representation of a static scene, achieving high-quality novel view synthesis. Consequently, some researchers have considered combining SfM and SLAM with NeRF, not only to reduce NeRF's dependence on the accuracy of input image poses but also to enhance the scene representation capability of SfM and SLAM. BARF~\cite{lin2021barf} was the first to integrate the core bundle adjustment (BA) algorithm from SfM with NeRF and adopted a coarse-to-fine reconstruction strategy, progressively aligning camera poses during the reconstruction process. To address the issue of camera poses being prone to local optima in BARF, L2G-NeRF~\cite{chen2023local} proposed a Local-to-Global alignment strategy, allowing camera poses to converge more easily to the global optimum. NoPe-NeRF~\cite{bian2023nope} introduced monocular depth information and key point matching information, respectively, to constrain the relative camera pose relationships, ensuring global consistency of camera poses. LocalRF~\cite{meuleman2023progressively} proposed a progressive strategy based on video sequences to gradually optimize local regions. CF-NeRF~\cite{yan2023cfnerf} employed an incremental learning approach to enable reconstruction under complex trajectories. LU-NeRF~\cite{cheng2023lu} introduced a Local-to-Global pose estimation strategy, enabling pose estimation and scene reconstruction from datasets with completely unknown camera poses. As for NeRF-based SLAM methods, iMAP~\cite{sucar2021imap} adopted two threads: Tracking and Mapping. The Tracking thread optimizes the camera pose using the current model and performs key frame selection, while the Mapping thread jointly optimizes the poses of the keyframes and the model. iMAP uses a single MLP to represent the entire scene, limiting its scalability. Nice-SLAM~\cite{zhu2022nice} improved upon iMAP by combining Hierarchical Feature Grids and MLP as the scene representation, enabling the application to large-scale scenes. Co-SLAM~\cite{wang2023co} and e-SLAM~\cite{johari2023eslam} introduced multi-resolution hash encoding and tri-plane representations, respectively, to improve system frame rate and scene representation capability, building upon Nice-SLAM. Regrettably, current NeRF-based SLAM methods necessitate dense image sequences, while NeRF-based SfM techniques face difficulties in accommodating complex camera trajectories. To tackle these limitations, we introduce ~\ourmethod, an incremental optimization framework that leverages additional correspondence constraints.~\ourmethod{} employs an incremental optimization process, iteratively refining the reconstruction as new images are integrated.

\noindent\textbf {Correspondence in pose-estimate} Local feature matching plays a crucial role in SfM and SLAM. The traditional feature matching pipeline consists of three main steps: feature detection, feature descriptor computation, and feature matching. Through feature detectors, the search space for matching can be effectively reduced, and the generated sparse matches are often sufficient to handle most tasks. However, in low-texture regions or repetitive patterns, these methods often fail due to the inability to detect sufficient feature points. With the flourishing development of deep learning, some researchers have started to leverage data-driven dense feature matching to enhance the accuracy and robustness of SfM and SLAM. For instance, Droid-SLAM~\cite{teed2021droid} utilizes the dense optical flow learned by RAFT~\cite{teed2020raft} for feature matching, achieving higher accuracy and robustness compared to traditional SLAM, and rarely failing in experimental scenarios. Detector-free SfM~\cite{he2023dfsfm} leverages the matching strategy of Loftr~\cite{sun2021loftr} without feature detectors, exhibiting significant advantages in low-texture regions and winning multiple competitions. Droid-SLAM and Detector-free SfM focus on pose estimation and reconstructing the scene with sparse points. On the other hand, SPARF~\cite{truong2023sparf} utilizes the correspondences from DKM matching for implicit neural reconstruction under noisy poses with several views. Different from these works, our method focuses on an incremental pose and implicit scene joint optimization pipeline with complex trajectories.

\section{Method}
\label{sec:method}

\begin{figure*}[!htbp]
     \centering
     \includegraphics[width=0.9\textwidth]{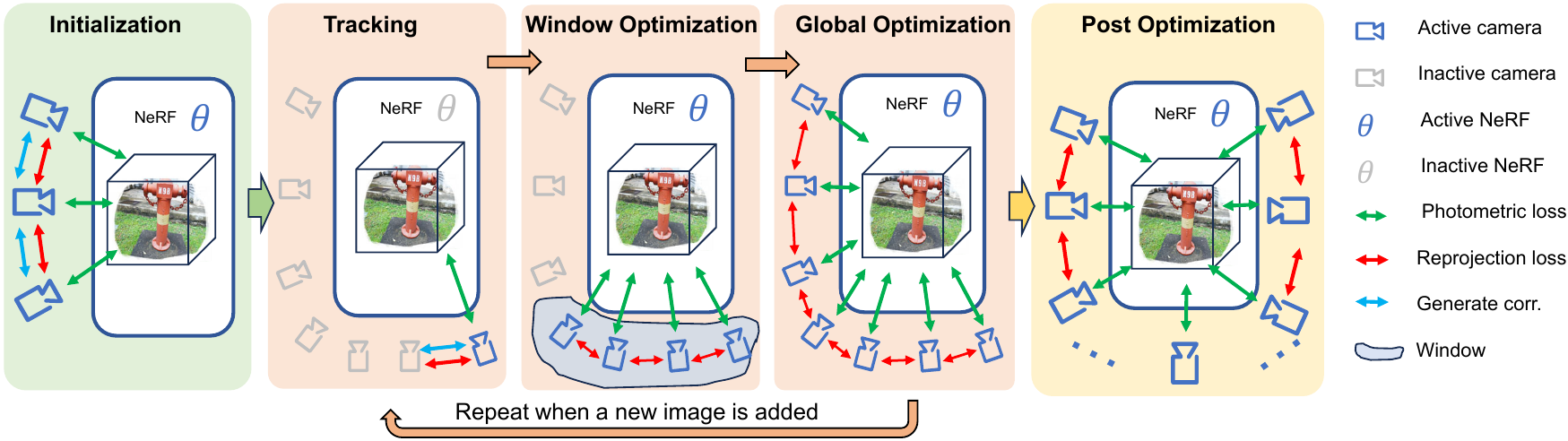}
     \caption{Our incremental optimization pipeline for neural radiance fields and pose estimation.}
\figlabel{method}
\end{figure*}

We first present the formulation of incremental scene reconstruction and pose estimation: given a set of sequential images $ \mathcal{I} = \{I_1, I_2,..., I_n\}$, where $n$ represents the $n^{th}$ frame captured in a camera trajectory, we aim to jointly optimize the poses $\mathcal{P} = \{P_1, P_2,..., P_n\}$ for the images and a neural radiance field model $\Theta$ representing the 3D scene captured by the images by adding one image at a time sequentially. 
To achieve the goal, we design an incremental reconstruction pipeline for the neural radiance field without pose priors. Our pipeline consists of five parts as shown in~\figref{method}. The scene is initialized using a small set of images. Subsequently, for each input image, tracking is applied to estimate the rough camera pose for a new image. A window optimization is followed to refine the poses of images within the window and also reconstruct the local structural components. To further incorporate the consistency of all visited camera poses, 
 global optimization (bundle adjustment) is performed on all images to optimize the global camera poses and the overall scene structure. Tracking, window, and global optimization repeat until all images are added. After all images are added, post optimization iteratively refines the entire scene and all camera poses until convergence.

\subsection{Preliminary: NeRF with Pose Optimization}
\seclabel{background}
We define the camera projection function $\pi$ that projects a space point $\mathbf{x} \in \mathcal{R}^3$ to a pixel $p \in \mathcal{R}^2$ as 
\begin{equation}
    p = \pi(\mathbf{x},P,K),
    \eqlabel{project}
\end{equation}
where camera pose $P$ is the camera-to-world transformation of image $I$ and $K$ is the intrinsic (we assume all images share the same intrinsic in a trajectory).  $P = [R,t]$, where $R \in SO(3)$ represents rotation and $t \in \mathcal{R}^3$ translation. The homogenization operations are omitted for clarity. The backprojection function $\pi^{-1}$ projects the pixel coordinate location $p$ into a space point with depth $z$
\begin{equation}
    x = \pi^{-1}(p,P,K,z).
    \eqlabel{backproject}
\end{equation}

NeRF maps a 3D location $\mathbf{x} \in \mathbb{R}^3$ and a view direction $\mathbf{d} \in \mathbb{R}^3$ to a radiance color $\mathbf{c} \in \mathbb{R}^3$ and volume density $\sigma \in \mathbb{R}$ with an MLP parameterized by $\Theta$. It  optimizes the model $\Theta$ and camera poses $\mathcal{P}$ by minimizing  photometric loss between rendered images $\mathcal{\hat{I}}$ and input images $\mathcal{I}$

\begin{equation}
\Theta^{*}, P^{*} = \mathop{\arg\min_{\Theta,P}}\mathcal{L}_{photo}(\hat{\mathcal{I}},\mathcal{P}\mid\mathcal{I}),
\eqlabel{barfmethod}
\end{equation}
\noindent where $\mathcal{L}_{photo} = \frac{1}{n}\sum_{1}^{n}||\hat{I_i}(P_i,\Theta)-I_i||_2^2$ and $\hat{I}_i$ can be obtained through volume rendering. For each pixel $q$ in image $\hat{I}_i$, its color is rendered by aggregating predicted colors $\mathbf{c}$ and densities $\sigma$ alone the ray $r = \mathbf{o} + \mathbf{d}s$ where $\mathbf{o}$ and $\mathbf{d}$ can be obtained by $\mathbf{o},\mathbf{d} = \pi^{-1}(p,P,K)$ and $s \in (s_{near},s_{far}]$ represents sample distance. 

\begin{equation}
\hat{I}_i(q) = \int_{s_{near}}^{s_{far}}T(s)\sigma(r(s))\mathbf{c}(r(s),\mathbf{d})ds,
\eqlabel{calccolor}
\end{equation}

\noindent where $T(s) = \exp(-\int_{s_{near}}^s\sigma(r(h))dh)$ indicates how much light is
transmitted on ray up to $s$. In the same way, depth map $\hat{D}_i$ can be rendered.

\begin{equation}
\hat{D}_i(q) = \int_{s_{near}}^{s_{far}}T(s)\sigma(r(s))sds.
\eqlabel{calcdepth}
\end{equation}

\subsection{Reprojected Geometric Image Distance}
\seclabel{rploss}

As aforementioned, we aim to incorporate the geometric distance constraint from correspondences between images to jointly optimize camera poses and  3D scene model parameters under complex trajectories. For correspondence generation, many existing correspondence learning networks can be exploited. In this work we choose DKM~\cite{edstedt2023dkm} as it is the current state-of-the-art work in dense correspondence matching, achieving high matching accuracy and strong robustness. Given a pair of adjacent images $I_1, I_2$ in the input trajectory $ \mathcal{I}$, a pre-trained DKM model can generate full-resolution pixel-level correspondences between them while predicting the confidence of each pixel match $\alpha$. We use a set $M$ to represent the output correspondences for two input images
\begin{small}
\begin{equation}
    M = \{m=(q,p,\alpha) \mid q\in I_1,p\in I_2,\alpha\in (0,1]\}.
\eqlabel{gencpd}
\end{equation}
\end{small}

According to the multiview geometry theory ~\cite{hartley2003multiple}, four pairs of correspondences can solve the relative pose between the images and using the triangulation technique, the 3D scene points for the pairs can be acquired. Though the correspondences only produce sparse 3D points and our representation is implicit, the theory provides important information for our problem: 1) correspondences provide a way to solve the pose estimation problem without knowing a 3D scene; 2) the pose estimation requires only sparse correspondences for the pose estimation; 3) the correspondences embed the 3D information. 

We define our geometric image distance for the pose estimation between two images under implicit scene representation as follows. As correspondence in practice often contains noise, we randomly sample $N_m$ correspondences from the correspondence set $M$ with confidence above a threshold $t_m$, to serve as a correspondences set $M_{sparse}$ for pose estimation, 
\begin{equation}
    M_{sparse} = \{m_1,m_2,...,m_{N_m}\} \sim \{m \mid m \in M, \alpha > t_m \}.
\eqlabel{selectcpd}
\end{equation}

In classic SfM pipelines, algorithms like RANSAC ~\cite{fischler1981random} are exploited to remove outliers and get robust poses and 3D points. Though differentiable RANSAC~\cite{10378404} may be deployed in our implicit joint optimization, we choose a simpler strategy: 1) weighting the reprojection error according to the confidence of correspondence estimation; 2) randomly sampling a small set of samples in each iteration.
In every training iteration for the scene model $\Theta$, $N_s$ pairs of correspondences are randomly fetched from $M_{sparse}$. Sampled correspondence sets with the confidence from multiple different iterations to optimize the pose and/or scene parameters serve as the similar purpose of RANSAC. 

For each correspondence $(q,p,\alpha)$, its depth value $z(q;\Theta)$ can be rendered by \eqref{calcdepth}, and $q$ can be backprojected into a 3D point and reprojected to $I_2$. Then we have the geometric image distance
\begin{small}
\begin{equation}
    \mathcal{L}_{rp}(I_1,I_2) = \frac{1}{N_s}\sum_1^{N_s}\alpha|\pi((\pi^{-1}(q,P_1,K,z(q;\Theta))),P_2,K)-p|,
    \eqlabel{rploss}
\end{equation}
\end{small}
which is reprojection error. $P_1,P_2$ are the poses to be estimated for images $I_1, I_2$ respectively.  The gradient with respect the camera poses  $P_1, P_2$, rendered depths and further the model $\Theta$ can be obtained from \eqref{rploss}, indicating the gradients can help the depth and scene reconstruction in addition to pose estimation.

The reprojection error is referred as \textit{Gold Standard}~\cite{hartley2003multiple}. The error is a quadratic convex function of pose parameters, pointing the right direction for pose optimization without local minima. However, the pixel value difference does not necessarily correlate to the correct direction for the pose optimization as shown in BARF~\cite{lin2021barf}. Compared with the pixel value difference, the reprojection error also provides more robust gradients for the scene estimation as the pixel value difference is susceptible to the lighting, occlusion, sparse views, etc, while the correspondence learning network for the reprojection error learns these factors during training.

\subsection{Tracking}
\seclabel{tracking}

When a new frame $I_i$ is added to the training process, tracking provides a rough estimate of the camera pose, which becomes particularly crucial when there exist violent changes in camera motion. The pose of the new frame is initialized based on the previous frame $P_i = P_{i-1}$. Tracking performs pose estimation through the reprojection error with adjacent frames and photometric loss with the scene. The loss function can be formulated as:

\begin{equation}
    \mathcal{L}(\mathcal{E}) = \mathcal{L}_{nrp}(\mathcal{E}) + \mathcal{L}_{photo}(\mathcal{E}),
    \eqlabel{loss}
\end{equation}

\begin{equation}
    \mathcal{L}_{nrp}(\mathcal{E}) = \frac{1}{N_e * 2 -2}\sum_1^{N_e}(\mathcal{L}_{rp}(I_i,I_{i-1}) + \mathcal{L}_{rp}(I_i,I_{i+1})),
    \eqlabel{nrp}
\end{equation}
where $\mathcal{L}_{nrp}$ is reprojection loss for paris of neighboring images and 
$\mathcal{E} = \{I_1,I_2,...,I_{N_e}\}$ is an optimization frame set  with $N_e$ frames. 

For the tracking, $\mathcal{E}_{tracking} = \{I_{i-1}, I_{i}\}$ consists of a new frame and the preceding frame ($N_e=2$) and we optimize only the pose of the newly added frame $P_i$, keeping other optimizable pose parameters and the network parameters $\Theta$ fixed.

\noindent{\textbf{Initialization}}  is crucial as it affects the subsequent tracking performance and the final pose estimation and reconstruction quality. In our approach, we select the first $N_{init}$ images in the sequence as the initialization images set $\mathcal{E}_{init} = \{I_1,I_2,...,I_{N_{init}}\}$. The initialization is achieved by minimizing \eqref{loss}. Due to the difficulty in optimizing the rotation $R$ parameters in the initial stages, we fix the rotation parameters and do not optimize them during the initialization.

\subsection{Joint Optimization}
\seclabel{joint}

Using the center-based graph and photo loss in equation \eqref{barfmethod}  for the joint optimization or bundle adjustment is susceptible to local minima when only RGB inputs are available. It only constrains all the poses in consistency with the scene while reconstructing the scene with implicit neural radiance fields from only RGB images (especially sparse images) are prone to converge to wrong geometry, which is demonstrated in many previous works \cite{deng2022depth,niemeyer2022regnerf,wang2023digging,song2024darf}. If the scene is stuck in a local minimum, the bundle adjustment can not get the pose right as long as the poses conform with the twisted scene. As shown in \tableref{tablebuster}, though the poses from BARF\cite{lin2021barf}, L2G-NeRF\cite{chen2023local}, Nope-NeRF\cite{bian2023nope} are fairly deviated, visual metrics like PSNR for the scene maintains high, causing the pose not able to escape the local minima. In contrast, we construct pose graphs with constraints between pose edges for the joint optimization.  The subgraphs formed between the poses and between the pose and the scene (shown in ~\figref{combine} (a) right) enable consistency of all the in-between poses in the image set. The reprojection error forces the consistency of the adjacent poses with the correspondences of input images. 
Further, we design a combination of local window and global BA strategy to balance the integration of new information and consistency with existing estimation.

Our joint optimization uses the same equation \eqref{nrp} in tracking. However, the scene model $\Theta$ is learnable, and different frame sets $\mathcal{E} $  for the optimization are maintained.

\noindent{\textbf{Window optimization}} To joint optimize the scene and the camera pose for a newly added image,  window optimization selects the most recent $N_{window}$ frames as the optimization set $\mathcal{E}_{window} = \{I_{i-N_{window}+1},..., I_{i-1}, I_{i}\}$. 
The window optimization process fixes the camera poses outside the window and optimizes the camera poses within the window and network $\Theta$ using \eqref{loss} by performing local bundle adjustment. Different from the existing implicit SfM and SLAM methods which consist of tracking for the newly added image and global bundle adjustment for all the input images, the extra window optimization improves the pose estimation accuracy by enhancing consistency with the near previous frames and makes the network learn faster from the information of the new frame by leaving older frames out.

\noindent{\textbf{Global optimization}} Relying solely on tracking and window optimization can lead to cumulative errors and even failure in pose estimation. To address this issue, global optimization incorporates all frames currently added to the training process into the optimization set $\mathcal{E}_{global} = \{I_1, I_2,..., I_i\}$. By applying \eqref{loss} to optimize all frames simultaneously, global optimization significantly enhances the robustness and accuracy of pose estimation.

\begin{figure*}[!htbp]
     \centering
     \includegraphics[width=0.9\textwidth]{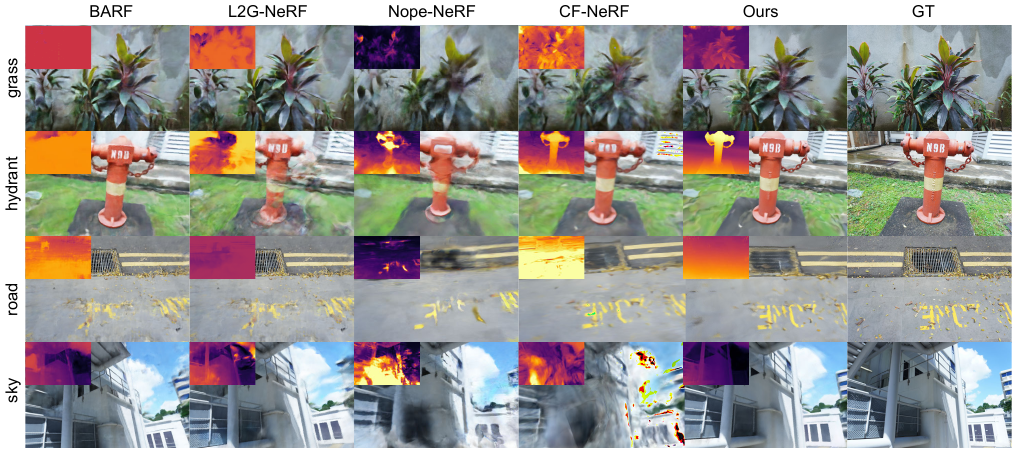}
     \caption{Qualitative Comparison on Free-Dataset~\cite{wang2023f2}. Rendered views and depths (top left corner of each image)}
\figlabel{rgbcmpfree}
\end{figure*}

\noindent{\textbf{Post optimization}} Before all frames are added, the learning rate of the network $lr_{\Theta}$ and poses $lr_{\hat{\mathcal{P}}}$ are fixed. The positional encoding control parameter $\alpha_{pe}$ is also fixed. After the whole frames are added, the post optimization process gradually reduces the learning rate and increases the frequency of positional encoding to iteratively refine the entire scene and camera poses through \eqref{loss}, ultimately obtaining the final results.

\subsection{Training procedure}
\seclabel{overall}

\noindent{\textbf{Training Pipeline}} The pipeline is initialized with $N_{init}$ frames. The stage is optimized for $\beta_{init}$ iterations. Afterwards, for each subsequent frame added, tracking is performed for $\beta_{tracking}$ iterations, window optimization for $\beta_{window}$ iterations, and global optimization for $\beta_{global}$ iterations. These three stages repeat with a new frame added and continue until all frames have been added. Finally, the post optimization stage consisting of $\beta_{post}$ iterations is conducted to further refine the reconstruction.

\noindent{\textbf{Positional Encoding}} The coarse-to-fine positional encoding plays an important role in accurate pose estimation~\cite{lin2021barf}, as excessively high frequencies can hinder this process. To address this, we employ the BARF~\cite{lin2021barf} positional encoding frequency control method. Specifically, before the post-optimization stage, we ensure a low-frequency setting for the positional encoding control parameter $\alpha_{pe}$. During post optimization, we keep the same coarse-to-fine strategy as the BARF.

\section{Experiments}

\subsection{Experiment Settings}

\noindent\textbf{Dataset} We evaluated our method on the challenging datasets with complex trajectories, \textbf{NeRFBuster}~\cite{warburg2023nerfbusters} and \textbf{Free-Dataset}~\cite{wang2023f2}. \textbf{NeRFBuster} consists of a total of 12 scenes, with most trajectories revolving around a central object. We employ sequences selected by CF-NERF~\cite{yan2023cfnerf}, with approximately 50 images per scene. We chose every 8th image from each sequence for novel view synthesis as the test set. All images are downsampled to a resolution of $480\times270$. Ground truth poses are estimated using COLMAP, as provided by CF-NeRF. \textbf{Free-Dataset} comprises 7 scenes with arbitrary trajectories, predominantly in outdoor environments characterized by highly dynamic camera motions. We select 50 images per scene in sequential order, and every 8th image is designated as the test set. The images are downsampled to a resolution of $312\times487$, and the ground truth poses are obtained through COLMAP~\cite{schonberger2016structure}.

\noindent\textbf{Implementation Details} Our approach is implemented based on the BARF~\cite{lin2021barf} framework. The majority of the hyperparameters in our network model align with the BARF Real-World Scenes setting, including the network learning rate $lr_{\Theta}$ decay from $1 \times 10^{-3}$ to $1 \times 10^{-4}$, pose learning rate $lr_{\mathcal{P}}$ decay from $3 \times 10^{-3}$ to $1 \times 10^{-5}$, inverse sampling of 128 points along each ray with an inverse range of $[1,0)$, a batch size of 1024, and linearly adjust $\alpha$ for post optimization phase from iteration 20K to 100K. We randomly select $N_m = 10000$ correspondences with confidence scores higher than $t_m=0.2$ from dense correspondences as sparse correspondences. During each iteration, $N_s=\frac{256}{len(\mathcal{E})}$ correspondences are randomly chosen from this set for reprojection loss. For NeRFBuster scenes, we set $N_{init} = 3$, $N_{window}= 4$, $\beta_{init}=2000$, $\beta_{tracking}=100$, $\beta_{window}=200$, $\beta_{global}=500$, and $\beta_{post} = 200K$. As the frames in in Free-Dataset exhibit larger camera motions and smaller overlap, the network requires more iterations to estimate poses accurately and achieve convergence.
For Free-Dataset scenes, we set $N_{init} = 3$, $N_{window}= 4$, $\beta_{init}=4000$, $\beta_{tracking}=200$, $\beta_{window}=400$, $\beta_{global}=900$, and $\beta_{post} = 200K$.

\begin{table*}[!ht]
    \caption{Evaluations of the pose accuracy (top 2 rows) and the novel view quality (bottom 3 rows) on NeRFBuster~\cite{warburg2023nerfbusters}. $\Delta T$ is the transition error in ground truth scale and  $\Delta R$ is rotation error in degree.}
    \tablelabel{tablebuster}
    \centering
    
    \setlength{\tabcolsep}{2mm}{
    \scriptsize
    
    \begin{tabular}{c|c|ccccccccccccc}
    \toprule
         Metrics & Method & aloe & art & car & century & garbage & flowers & picnic & pikachu & pipe & plant & roses & table & mean \\ \midrule
         \multirow{5}{*}{$\Delta R$↓} & BARF~\cite{lin2021barf} & 128.599 & 45.739 & 97.148 & 114.261 & 103.259 & 79.702 & 81.418 & 112.996 & 166.701 & 140.270 & 125.974 & 139.675 & 111.312 \\
         & L2G-NeRF~\cite{chen2023local} & 117.475 & 28.247 & 161.862 & 59.730 & 90.889 & 88.750 & 99.076 & 123.543 & 106.838 & 71.551 & 139.057 & 144.848 & 102.656\\
         & Nope-NeRF~\cite{bian2023nope} & 101.589 & 32.345 & 113.063 & 150.253 & 149.459 & 161.859 & 148.710 & 158.059 & 99.836 & 138.816 & 150.050 & 114.783 & 126.569\\
         & CF-NeRF~\cite{yan2023cfnerf} &  6.703&  76.306&  29.079&  11.013&  74.163&  10.672&  109.868&  13.243&  122.345&  18.664&  3.903 &  3.835 & 39.983\\
         & Ours & \textbf{3.163} & \textbf{3.151} & \textbf{0.701} & \textbf{2.343} & \textbf{0.902} & \textbf{0.481} & \textbf{1.938} & \textbf{7.708} & \textbf{2.302} & \textbf{6.302} & \textbf{0.570} & \textbf{1.154} & \textbf{2.560}\\
         \midrule
         \multirow{5}{*}{$\Delta T$↓} & BARF & 6.039 & 4.040 & 5.043 & 5.434 & 4.663 & 4.693 & 3.007 & 3.772 & 3.763 & 5.865 & 4.952 & 4.076 & 4.612 \\
         & L2G-NeRF & 4.986 & 4.402 & 4.764 & 5.895 & 4.272 & 4.926 & 4.214 & 6.451 & 5.592 & 2.764 & 5.055 & 4.199 & 4.795 \\
         & Nope-NeRF& 5.151 & 5.302 & 5.401 & 3.202 & 5.571 & 4.742 & 4.819 & 3.757 & 4.983 & 5.896 & 5.399 & 5.817 & 5.004\\
         & CF-NeRF& 0.637 & 1.549 & 1.621 & 0.497 & 0.548 & 0.745 & 1.285 &  0.879 &  5.757 &  0.685 &  0.182 & 0.274 & 1.222\\
         & Ours &\textbf{0.168} & \textbf{0.030} & \textbf{0.035} & \textbf{0.134} & \textbf{0.039} & \textbf{0.039} & \textbf{0.106} & \textbf{0.548} & \textbf{0.164} & \textbf{0.225} & \textbf{0.038} & \textbf{0.045} & \textbf{0.131} \\
         \midrule
         \multirow{5}{*}{$PSNR$↑} & BARF & 23.56 & 20.55 & 23.69 & 18.73 & 19.92 & 23.14 & 22.91 & 31.58 & \textbf{23.43} & 29.38 & 21.87 & 25.88 & 23.72\\
         & L2G-NeRF & 23.47 & 22.58 & 23.98 & 19.32 & 20.36 & 24.52 & 22.18 & \textbf{33.66} & 21.99 & \textbf{29.63} & 21.74 & 25.60 & 24.09\\
         & Nope-NeRF& 22.42 & 21.53 & 22.62 & 19.55 & 20.59 & 21.91 & 22.97 & 27.39 & 21.33 & 25.69 & 19.91 & 26.83 & 22.73\\
         & CF-NeRF& 23.32 & 23.50 & 22.04 & 21.35 & 21.30 & 23.91 & \textbf{23.31} & 31.51 & 22.24 & 25.89 & 23.42 & 26.71 & 24.04\\
         & Ours neighbor& 23.17&  25.76&  24.90&  21.64&  21.62&  26.14&  23.04 &  30.50&  23.02&  27.19&  22.14& 30.85 & 25.00\\
         & Ours sim(3) & \textbf{24.36} & \textbf{26.73} & \textbf{27.41} & \textbf{22.56} & \textbf{22.69} & \textbf{27.37} & 23.04 & 22.91 & 23.13 & 22.64 & \textbf{29.63} & \textbf{32.73} & \textbf{25.43}\\
         \midrule
         \multirow{5}{*}{$SSIM$↑} & BARF & 0.59 & 0.68 & 0.72 & 0.52 & 0.54 & 0.72 & \textbf{0.54} & 0.92 & \textbf{0.62} & 0.85 & 0.69 & 0.84 & 0.69 \\
         & L2G-NeRF & 0.58 & 0.74 & 0.74 & 0.56 & 0.53 & 0.76 & 0.50 & \textbf{0.94} & 0.55 & \textbf{0.87} & 0.69 & 0.83 & 0.69 \\
         & Nope-NeRF& 0.52 & 0.73 & 0.69 & 0.57 & 0.56 & 0.67 & 0.53 & 0.88 & 0.53 & 0.80 & 0.64 & 0.86 & 0.67\\
         & CF-NeRF&  0.56&  0.74&  0.66&  0.63&  0.57&  0.72&  0.53& 0.93 & 0.54
         & 0.80 & 0.72  & 0.85 & 0.69\\
         & Ours neighbor& 0.56& 0.82& 0.74&  0.63& 0.58&  0.78&  0.52&  0.91&  0.58& 0.83& 0.71&  0.89  & 0.71\\
         & Ours sim(3) & \textbf{0.61} & \textbf{0.83} & \textbf{0.79} & \textbf{0.65} & \textbf{0.63} & \textbf{0.81} & 0.52 & 0.76 & 0.59 & 0.71 & \textbf{0.88} & \textbf{0.91} & \textbf{0.72} \\
         \midrule
         \multirow{5}{*}{$LPIPS$↓} & BARF & 0.36 & 0.20 & 0.29 & 0.40 & \textbf{0.38} & 0.27 & \textbf{0.44} & \textbf{0.09} & \textbf{0.33} & \textbf{0.14} & 0.24 & 0.23 & \textbf{0.28} \\
         & L2G-NeRF & 0.37 & 0.18 & 0.28 & \textbf{0.35} & 0.42 & \textbf{0.14} & 0.49 & 0.10 & 0.39 & \textbf{0.14} & 0.24 & 0.24 & 0.29 \\
         & Nope-NeRF& 0.51 & 0.33 & 0.43 & 0.48 & 0.50 & 0.47 & 0.54 & 0.30 & 0.52 & 0.34 & 0.48 & 0.35 & 0.44\\
         & CF-NeRF& 0.43 & 0.24 & 0.41 & 0.37 & 0.46 & 0.33 & 0.55 & 0.11 & 0.50 & 0.20 & 0.21 & 0.23 & 0.34\\
         & Ours neighbor& 0.36 & \textbf{0.14} & 0.26 & 0.50 & 0.44 & 0.22 & 0.49 & 0.11 & 0.41 & 0.22  & 0.14 & \textbf{0.17} & 0.29 \\
         & Ours sim(3) & \textbf{0.35} & \textbf{0.14} & \textbf{0.25} & 0.50 & 0.43 & 0.22 & 0.49 & 0.29 & 0.40 & 0.30 & \textbf{0.09} & \textbf{0.17} & 0.30 \\\bottomrule
    \end{tabular}
    }


\end{table*}

\noindent\textbf{Metrics} We primarily evaluate our method by assessing the quality of novel view synthesis and the accuracy of pose estimation. As the variables for the scene and the cameras are up to a 3D similarity transformation, existing work~\cite{lin2021barf, chen2023local,bian2023nope} aligns the optimized poses to the ground truth using Sim(3) with Procrusters analysis on the camera locations for pose error computation and test pose initialization  (termed as Sim(3) below) and then runs an additional test-time optimization on the trained model to reduce the pose error that may influence the view synthesis quality.  
Since all existing methods compared below except ours struggle to obtain reasonable initial test poses through Sim(3) and fail to perform the test-time optimization under the complex trajectories, we adopt the approach of Nope-NeRF~\cite{bian2023nope} for these methods, \ie initializing a test image pose with the estimated pose of the training frame that is closest to it (termed as neighbor below). For our method, we provide results using both initialization methods. We report PSNR, SSIM, and LPIPS for the view synthesis, and rotation and translation errors for the pose estimation.

\subsection{Comparison with Pose-Unknown Methods}

We compare with the state-of-the-art methods for joint optimization of scenes and poses from RGB images, \ie BARF~\cite{lin2021barf}, L2G-NeRF~\cite{chen2023local}, Nope-NeRF~\cite{bian2023nope}, and CF-NeRF~\cite{yan2023cfnerf}. 

\begin{figure*}[!htbp]
     \centering
     \includegraphics[width=0.9\textwidth]{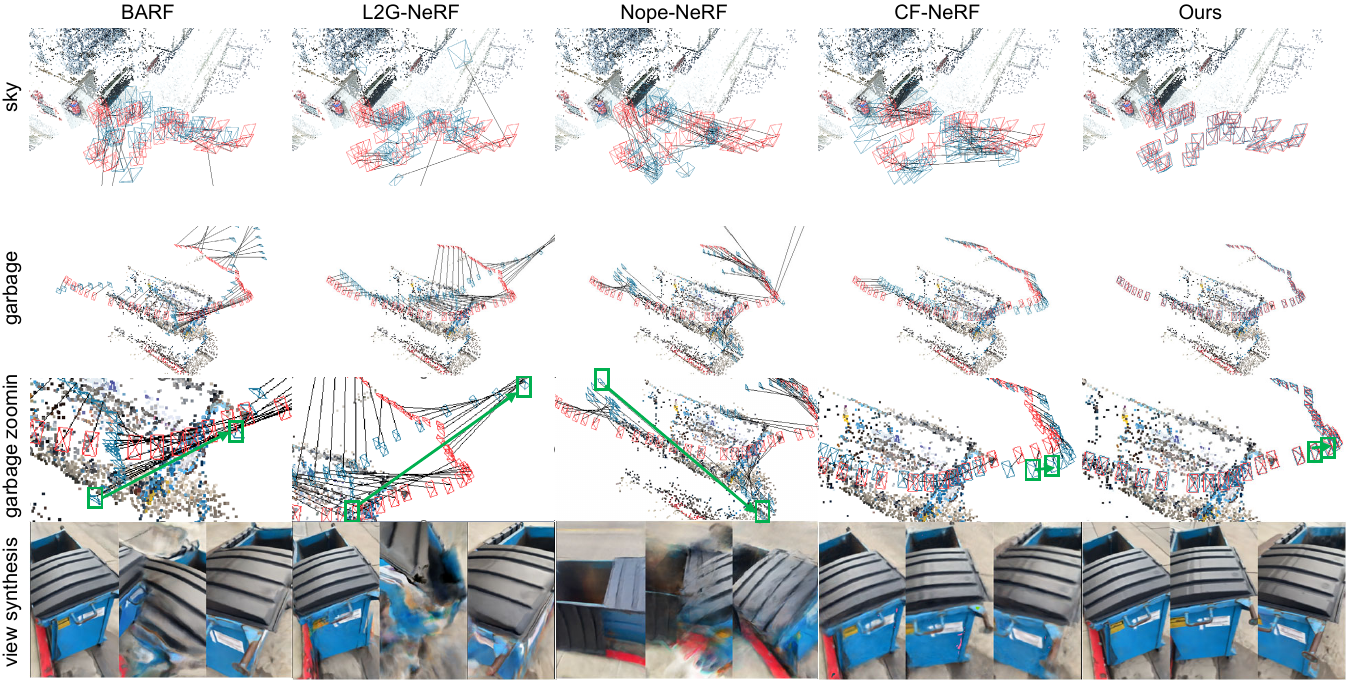}
     \caption{Trajectory comparison. We visualize camera poses of both estimated (blue) and COLMAP (red). Sparse 3D points for the scenes are from COLMAP. While there are abrupt changes in the trajectories of BARF, L2G-NeRF, and Nope-NeRF, the changes are steady along the trajectories of CF-NeRF and ours. The bottom row shows rendered interframes between two frames of abrupt changes denoted by green rectangles.}
\figlabel{cmptraj}
\end{figure*}

\begin{figure}[!htbp]
     \centering
     \includegraphics[width=\linewidth]{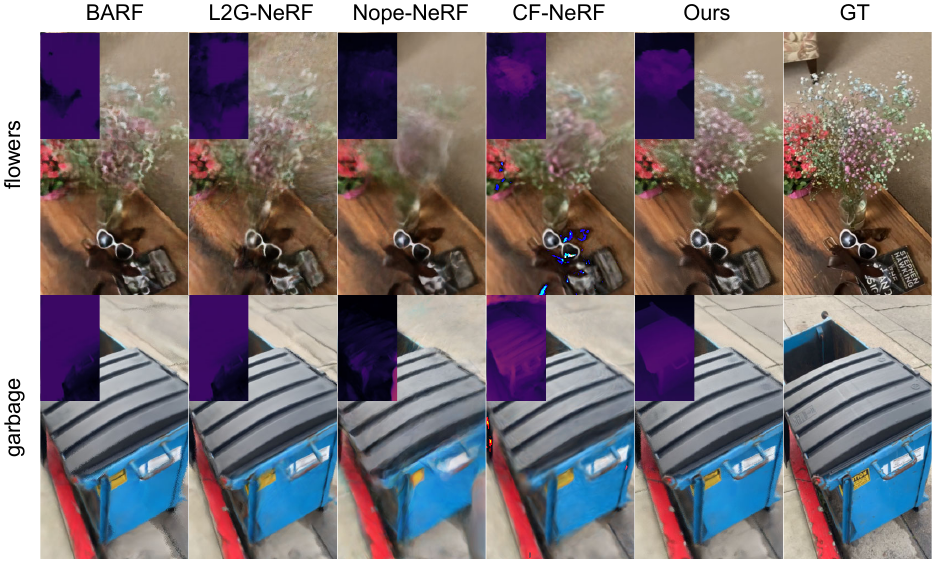}
     \caption{Qualitative Comparison on NeRFbuster~\cite{warburg2023nerfbusters}. Rendered views and depths (top left corner of each image).}
    \figlabel{cmprgbbuster}
\end{figure}

\noindent\textbf{Results on the Object Centered Dataset} \tableref{tablebuster} presents the pose evaluation results on the \textbf{NeRFBuster} dataset. BARF, L2G-NeRF, and Nope-NeRF initialize all poses as identity matrices and then perform bundle adjustment to jointly optimize poses and the scene. With pose initialization far from the actual poses and sparse views not able to effectively constrain the scenes, these methods frequently fail to recover the camera poses and geometry. CF-NeRF manages to estimate poses but still suffers from larger errors as CF-NeRF only constrains poses through photo loss.

Our method achieves significantly smaller errors compared to these methods. 

Despite the significant pose errors and poor structural quality of methods such as BARF, they still achieve surprisingly "good" results on PSNR, SSIM, and LPIPS  in \tableref{tablebuster}, almost on-par with our results on the view synthesis. 
We attribute this to overfitting both in the training stage and test-time optimization. In \figref{cmptraj}, we visualize the trajectory of the NeRFbuster \textbf{garbage} scene and select two frames with an abrupt change in the estimated camera trajectory. We then render novel views by interpolating between these poses. As shown in \figref{cmptraj} (b), the rendering results are unreasonable, while CF-NeRF and our method can render smooth view transitions (In the supplementary video, we further show BARF, L2G-NeRF, Nope-NeRF, and CF-NeRF renders view inconsistent effects). The other evidence is that these methods struggle to reconstruct the geometry as shown \figref{cmprgbbuster} and \figref{rgbcmpfree}. 
During test-time optimization, as the estimated trajectories of these methods diverge far from the ground truth and they fail to perform test-time optimization using Sim(3), the poses of the neighboring frames are used to initialize the test frames. This initialization causes the pose of a test frame to converge to a "pseudo " pose close to the estimated pose of the closest neighboring train frame. Then the network for the scene further overfits to the "pseudo" ground truth test pose and image ($P, I$ for example) pairs and renders an image  $I'$ using $P$ to calculate visual metrics with $I$.

\noindent\textbf{Results on the Free Trajectory Dataset} We also conduct our experiments on the \textbf{Free-Dataset}, which consists of more challenging scenarios with arbitrary trajectory variations and reduced frame overlap (please refer to the supplementary material for visualization of the sequences). In \tableref{tablefj}, we report the results on the Free-Dataset, which demonstrates that our method exhibits more significant advantages under more challenging scenes. \figref{rgbcmpfree} shows that in addition to the superior quality of the novel view synthesis, our method can produce depths of good quality while most existing methods fail to.

\begin{table}[!ht]
    \caption{Evaluations of the pose accuracy and the novel view quality on Free-Dataset~\cite{wang2023f2}. $\Delta T$ is the transition error in ground truth scale and  $\Delta R$ is rotation error in degree.}
    \tablelabel{tablefj}
    \centering
    \scriptsize
    
    \begin{tabular}{l|ccccc}
    \toprule
         
          Method & $\Delta R$↓ & $\Delta T$↓ & $PSNR$↑ & $SSIM$↑ & $LPIPS$↓   \\ \midrule
         BARF & 61.098 & 3.498 & 19.56 & 0.52 & 0.45   \\
         L2G-NeRF & 110.303 & 6.587 & 19.95 & 0.54 & 0.45   \\
         Nope-NeRF& 144.202 & 4.693 & 18.67 & 0.51 & 0.66   \\
         CF-NeRF&  55.329& 2.385 & 18.30 & 0.42 &  0.72  \\
         Ours neighbor & \textbf{2.805}  & \textbf{0.161} & 18.69 & 0.49 & 0.49    \\
         Ours sim(3) & \textbf{2.805} & \textbf{0.161} & \textbf{22.46} & \textbf{0.59} & \textbf{0.43}    \\
         \bottomrule
    \end{tabular}
    
\end{table}

\subsection{Ablation Study}
In this subsection, we conduct ablation studies to investigate the impact of various components in our method. We ablate \textbf{projection loss}, \textbf{tracking}, \textbf{window optimization}, and \textbf{global optimization} components individually. \tableref{components} shows that removing tracking, window, or global optimization leads to performance degradation, but the method remains functional. However, removing reprojection loss leads to dramatic pose errors. We refer readers to supplementary material for more results for the ablation.

\begin{table}[!ht]
    \caption{Ablation study on reprojection loss, tracking, window optimization, and global optimization.}
    \tablelabel{components}
    \centering
    \scriptsize
    \begin{tabular}{l|ccccc}
    \toprule
         
          Method & $\Delta R$↓ & $\Delta T$↓ & $PSNR$↑ & $SSIM$↑ & $LPIPS$↓   \\ \midrule
         Ours w/o reproj. loss & 56.040 & 1.904 & 16.18 & 0.44 & 0.63   \\          
         Ours w/o tracking & 5.302 & 0.280 & 23.66 & 0.67 & 0.35   \\
         Ours w/o window opt. & 3.562 & 0.182 & 24.57 & 0.70 & 0.33   \\
         Ours w/o global opt. & 6.189 & 0.234 & 22.66 & 0.64 & 0.40   \\
         Ours & \textbf{2.560} & \textbf{0.131} & \textbf{25.43} & \textbf{0.72} & \textbf{0.30}    \\
         \bottomrule
    \end{tabular}
\end{table}

\section{Conclusion}
We present ~\ourmethod, a method capable of recovering poses and reconstructing scenes from image sequences captured along complex trajectories. We first introduce correspondence and reprojected geometric image distance to impose extra constraints on the optimization graph, enabling robust and accurate pose estimation and scene structure reconstruction. Subsequently, we detail our incremental learning process for pose recovery, including initialization, tracking, window optimization, and global optimization. Through comparative and ablation experiments, we demonstrate the superiority of our method and the necessity of its individual components. 
Although our method enables joint pose estimation and reconstruction under complex camera trajectories, we only explore simple pose graphs. More sophisticated graph optimization is required for very long trajectories. Also, evaluation datasets, protocols, and metrics are required for complex camera trajectories as discussed in the paper, the current visual metrics can not fully reflect the reconstruction quality.

\par \vfill
\clearpage

\bibliographystyle{ACM-Reference-Format}
\bibliography{main}


\begin{thebibliography}{42}


\ifx \showCODEN    \undefined \def \showCODEN     #1{\unskip}     \fi
\ifx \showDOI      \undefined \def \showDOI       #1{#1}\fi
\ifx \showISBNx    \undefined \def \showISBNx     #1{\unskip}     \fi
\ifx \showISBNxiii \undefined \def \showISBNxiii  #1{\unskip}     \fi
\ifx \showISSN     \undefined \def \showISSN      #1{\unskip}     \fi
\ifx \showLCCN     \undefined \def \showLCCN      #1{\unskip}     \fi
\ifx \shownote     \undefined \def \shownote      #1{#1}          \fi
\ifx \showarticletitle \undefined \def \showarticletitle #1{#1}   \fi
\ifx \showURL      \undefined \def \showURL       {\relax}        \fi
\providecommand\bibfield[2]{#2}
\providecommand\bibinfo[2]{#2}
\providecommand\natexlab[1]{#1}
\providecommand\showeprint[2][]{arXiv:#2}

\bibitem[Abdelrasoul et~al\mbox{.}(2016)]%
        {abdelrasoul2016quantitative}
\bibfield{author}{\bibinfo{person}{Yassin Abdelrasoul}, \bibinfo{person}{Abu Bakar Sayuti~HM Saman}, {and} \bibinfo{person}{Patrick Sebastian}.} \bibinfo{year}{2016}\natexlab{}.
\newblock \showarticletitle{A quantitative study of tuning ROS gmapping parameters and their effect on performing indoor 2D SLAM}. In \bibinfo{booktitle}{\emph{2016 2nd IEEE international symposium on robotics and manufacturing automation (ROMA)}}. IEEE, \bibinfo{pages}{1--6}.
\newblock


\bibitem[Bailey et~al\mbox{.}(2006)]%
        {bailey2006consistency}
\bibfield{author}{\bibinfo{person}{Tim Bailey}, \bibinfo{person}{Juan Nieto}, \bibinfo{person}{Jose Guivant}, \bibinfo{person}{Michael Stevens}, {and} \bibinfo{person}{Eduardo Nebot}.} \bibinfo{year}{2006}\natexlab{}.
\newblock \showarticletitle{Consistency of the EKF-SLAM algorithm}. In \bibinfo{booktitle}{\emph{2006 IEEE/RSJ International Conference on Intelligent Robots and Systems}}. IEEE, \bibinfo{pages}{3562--3568}.
\newblock


\bibitem[Bian et~al\mbox{.}(2023)]%
        {bian2023nope}
\bibfield{author}{\bibinfo{person}{Wenjing Bian}, \bibinfo{person}{Zirui Wang}, \bibinfo{person}{Kejie Li}, \bibinfo{person}{Jia-Wang Bian}, {and} \bibinfo{person}{Victor~Adrian Prisacariu}.} \bibinfo{year}{2023}\natexlab{}.
\newblock \showarticletitle{Nope-nerf: Optimising neural radiance field with no pose prior}. In \bibinfo{booktitle}{\emph{Proceedings of the IEEE/CVF Conference on Computer Vision and Pattern Recognition}}. \bibinfo{pages}{4160--4169}.
\newblock


\bibitem[Campos et~al\mbox{.}(2021)]%
        {campos2021orb}
\bibfield{author}{\bibinfo{person}{Carlos Campos}, \bibinfo{person}{Richard Elvira}, \bibinfo{person}{Juan J~G{\'o}mez Rodr{\'\i}guez}, \bibinfo{person}{Jos{\'e}~MM Montiel}, {and} \bibinfo{person}{Juan~D Tard{\'o}s}.} \bibinfo{year}{2021}\natexlab{}.
\newblock \showarticletitle{Orb-slam3: An accurate open-source library for visual, visual--inertial, and multimap slam}.
\newblock \bibinfo{journal}{\emph{IEEE Transactions on Robotics}} \bibinfo{volume}{37}, \bibinfo{number}{6} (\bibinfo{year}{2021}), \bibinfo{pages}{1874--1890}.
\newblock


\bibitem[Castellanos et~al\mbox{.}(2007)]%
        {castellanos2007robocentric}
\bibfield{author}{\bibinfo{person}{Jos{\'e}~A Castellanos}, \bibinfo{person}{Ruben Martinez-Cantin}, \bibinfo{person}{Juan~D Tard{\'o}s}, {and} \bibinfo{person}{Jos{\'e} Neira}.} \bibinfo{year}{2007}\natexlab{}.
\newblock \showarticletitle{Robocentric map joining: Improving the consistency of EKF-SLAM}.
\newblock \bibinfo{journal}{\emph{Robotics and autonomous systems}} \bibinfo{volume}{55}, \bibinfo{number}{1} (\bibinfo{year}{2007}), \bibinfo{pages}{21--29}.
\newblock


\bibitem[Chen et~al\mbox{.}(2023)]%
        {chen2023local}
\bibfield{author}{\bibinfo{person}{Yue Chen}, \bibinfo{person}{Xingyu Chen}, \bibinfo{person}{Xuan Wang}, \bibinfo{person}{Qi Zhang}, \bibinfo{person}{Yu Guo}, \bibinfo{person}{Ying Shan}, {and} \bibinfo{person}{Fei Wang}.} \bibinfo{year}{2023}\natexlab{}.
\newblock \showarticletitle{Local-to-global registration for bundle-adjusting neural radiance fields}. In \bibinfo{booktitle}{\emph{Proceedings of the IEEE/CVF Conference on Computer Vision and Pattern Recognition}}. \bibinfo{pages}{8264--8273}.
\newblock


\bibitem[Cheng et~al\mbox{.}(2023)]%
        {cheng2023lu}
\bibfield{author}{\bibinfo{person}{Zezhou Cheng}, \bibinfo{person}{Carlos Esteves}, \bibinfo{person}{Varun Jampani}, \bibinfo{person}{Abhishek Kar}, \bibinfo{person}{Subhransu Maji}, {and} \bibinfo{person}{Ameesh Makadia}.} \bibinfo{year}{2023}\natexlab{}.
\newblock \showarticletitle{LU-NeRF: Scene and pose estimation by synchronizing local unposed nerfs}. In \bibinfo{booktitle}{\emph{Proceedings of the IEEE/CVF International Conference on Computer Vision}}. \bibinfo{pages}{18312--18321}.
\newblock


\bibitem[Chng et~al\mbox{.}(2022)]%
        {chng2022gaussian}
\bibfield{author}{\bibinfo{person}{Shin-Fang Chng}, \bibinfo{person}{Sameera Ramasinghe}, \bibinfo{person}{Jamie Sherrah}, {and} \bibinfo{person}{Simon Lucey}.} \bibinfo{year}{2022}\natexlab{}.
\newblock \showarticletitle{Gaussian activated neural radiance fields for high fidelity reconstruction and pose estimation}. In \bibinfo{booktitle}{\emph{European Conference on Computer Vision}}. Springer, \bibinfo{pages}{264--280}.
\newblock


\bibitem[Cui and Tan(2015)]%
        {cui2015global}
\bibfield{author}{\bibinfo{person}{Zhaopeng Cui} {and} \bibinfo{person}{Ping Tan}.} \bibinfo{year}{2015}\natexlab{}.
\newblock \showarticletitle{Global structure-from-motion by similarity averaging}. In \bibinfo{booktitle}{\emph{Proceedings of the IEEE International Conference on Computer Vision}}. \bibinfo{pages}{864--872}.
\newblock


\bibitem[Deng et~al\mbox{.}(2022)]%
        {deng2022depth}
\bibfield{author}{\bibinfo{person}{Kangle Deng}, \bibinfo{person}{Andrew Liu}, \bibinfo{person}{Jun-Yan Zhu}, {and} \bibinfo{person}{Deva Ramanan}.} \bibinfo{year}{2022}\natexlab{}.
\newblock \showarticletitle{Depth-supervised nerf: Fewer views and faster training for free}. In \bibinfo{booktitle}{\emph{Proceedings of the IEEE/CVF Conference on Computer Vision and Pattern Recognition}}. \bibinfo{pages}{12882--12891}.
\newblock


\bibitem[Edstedt et~al\mbox{.}(2023)]%
        {edstedt2023dkm}
\bibfield{author}{\bibinfo{person}{Johan Edstedt}, \bibinfo{person}{Ioannis Athanasiadis}, \bibinfo{person}{M{\aa}rten Wadenb{\"a}ck}, {and} \bibinfo{person}{Michael Felsberg}.} \bibinfo{year}{2023}\natexlab{}.
\newblock \showarticletitle{DKM: Dense kernelized feature matching for geometry estimation}. In \bibinfo{booktitle}{\emph{Proceedings of the IEEE/CVF Conference on Computer Vision and Pattern Recognition}}. \bibinfo{pages}{17765--17775}.
\newblock


\bibitem[Engel et~al\mbox{.}(2014)]%
        {engel2014lsd}
\bibfield{author}{\bibinfo{person}{Jakob Engel}, \bibinfo{person}{Thomas Sch{\"o}ps}, {and} \bibinfo{person}{Daniel Cremers}.} \bibinfo{year}{2014}\natexlab{}.
\newblock \showarticletitle{LSD-SLAM: Large-scale direct monocular SLAM}. In \bibinfo{booktitle}{\emph{European conference on computer vision}}. Springer, \bibinfo{pages}{834--849}.
\newblock


\bibitem[Fischler and Bolles(1981)]%
        {fischler1981random}
\bibfield{author}{\bibinfo{person}{Martin~A Fischler} {and} \bibinfo{person}{Robert~C Bolles}.} \bibinfo{year}{1981}\natexlab{}.
\newblock \showarticletitle{Random sample consensus: a paradigm for model fitting with applications to image analysis and automated cartography}.
\newblock \bibinfo{journal}{\emph{Commun. ACM}} \bibinfo{volume}{24}, \bibinfo{number}{6} (\bibinfo{year}{1981}), \bibinfo{pages}{381--395}.
\newblock


\bibitem[Gherardi et~al\mbox{.}(2010)]%
        {gherardi2010improving}
\bibfield{author}{\bibinfo{person}{Riccardo Gherardi}, \bibinfo{person}{Michela Farenzena}, {and} \bibinfo{person}{Andrea Fusiello}.} \bibinfo{year}{2010}\natexlab{}.
\newblock \showarticletitle{Improving the efficiency of hierarchical structure-and-motion}. In \bibinfo{booktitle}{\emph{2010 IEEE computer society conference on computer vision and pattern recognition}}. IEEE, \bibinfo{pages}{1594--1600}.
\newblock


\bibitem[Hartley and Zisserman(2003)]%
        {hartley2003multiple}
\bibfield{author}{\bibinfo{person}{Richard Hartley} {and} \bibinfo{person}{Andrew Zisserman}.} \bibinfo{year}{2003}\natexlab{}.
\newblock \bibinfo{booktitle}{\emph{Multiple view geometry in computer vision}}.
\newblock \bibinfo{publisher}{Cambridge university press}.
\newblock


\bibitem[He et~al\mbox{.}(2023)]%
        {he2023dfsfm}
\bibfield{author}{\bibinfo{person}{Xingyi He}, \bibinfo{person}{Jiaming Sun}, \bibinfo{person}{Yifan Wang}, \bibinfo{person}{Sida Peng}, \bibinfo{person}{Qixing Huang}, \bibinfo{person}{Hujun Bao}, {and} \bibinfo{person}{Xiaowei Zhou}.} \bibinfo{year}{2023}\natexlab{}.
\newblock \showarticletitle{Detector-Free Structure from Motion}. In \bibinfo{booktitle}{\emph{arxiv}}.
\newblock


\bibitem[Jeong et~al\mbox{.}(2021)]%
        {jeong2021self}
\bibfield{author}{\bibinfo{person}{Yoonwoo Jeong}, \bibinfo{person}{Seokjun Ahn}, \bibinfo{person}{Christopher Choy}, \bibinfo{person}{Anima Anandkumar}, \bibinfo{person}{Minsu Cho}, {and} \bibinfo{person}{Jaesik Park}.} \bibinfo{year}{2021}\natexlab{}.
\newblock \showarticletitle{Self-calibrating neural radiance fields}. In \bibinfo{booktitle}{\emph{Proceedings of the IEEE/CVF International Conference on Computer Vision}}. \bibinfo{pages}{5846--5854}.
\newblock


\bibitem[Jiang et~al\mbox{.}(2013)]%
        {jiang2013global}
\bibfield{author}{\bibinfo{person}{Nianjuan Jiang}, \bibinfo{person}{Zhaopeng Cui}, {and} \bibinfo{person}{Ping Tan}.} \bibinfo{year}{2013}\natexlab{}.
\newblock \showarticletitle{A global linear method for camera pose registration}. In \bibinfo{booktitle}{\emph{Proceedings of the IEEE international conference on computer vision}}. \bibinfo{pages}{481--488}.
\newblock


\bibitem[Johari et~al\mbox{.}(2023)]%
        {johari2023eslam}
\bibfield{author}{\bibinfo{person}{Mohammad~Mahdi Johari}, \bibinfo{person}{Camilla Carta}, {and} \bibinfo{person}{Fran{\c{c}}ois Fleuret}.} \bibinfo{year}{2023}\natexlab{}.
\newblock \showarticletitle{Eslam: Efficient dense slam system based on hybrid representation of signed distance fields}. In \bibinfo{booktitle}{\emph{Proceedings of the IEEE/CVF Conference on Computer Vision and Pattern Recognition}}. \bibinfo{pages}{17408--17419}.
\newblock


\bibitem[Lin et~al\mbox{.}(2021)]%
        {lin2021barf}
\bibfield{author}{\bibinfo{person}{Chen-Hsuan Lin}, \bibinfo{person}{Wei-Chiu Ma}, \bibinfo{person}{Antonio Torralba}, {and} \bibinfo{person}{Simon Lucey}.} \bibinfo{year}{2021}\natexlab{}.
\newblock \showarticletitle{Barf: Bundle-adjusting neural radiance fields}. In \bibinfo{booktitle}{\emph{Proceedings of the IEEE/CVF International Conference on Computer Vision}}. \bibinfo{pages}{5741--5751}.
\newblock


\bibitem[Meuleman et~al\mbox{.}(2023)]%
        {meuleman2023progressively}
\bibfield{author}{\bibinfo{person}{Andreas Meuleman}, \bibinfo{person}{Yu-Lun Liu}, \bibinfo{person}{Chen Gao}, \bibinfo{person}{Jia-Bin Huang}, \bibinfo{person}{Changil Kim}, \bibinfo{person}{Min~H Kim}, {and} \bibinfo{person}{Johannes Kopf}.} \bibinfo{year}{2023}\natexlab{}.
\newblock \showarticletitle{Progressively optimized local radiance fields for robust view synthesis}. In \bibinfo{booktitle}{\emph{Proceedings of the IEEE/CVF Conference on Computer Vision and Pattern Recognition}}. \bibinfo{pages}{16539--16548}.
\newblock


\bibitem[Mildenhall et~al\mbox{.}(2021)]%
        {mildenhall2021nerf}
\bibfield{author}{\bibinfo{person}{Ben Mildenhall}, \bibinfo{person}{Pratul~P Srinivasan}, \bibinfo{person}{Matthew Tancik}, \bibinfo{person}{Jonathan~T Barron}, \bibinfo{person}{Ravi Ramamoorthi}, {and} \bibinfo{person}{Ren Ng}.} \bibinfo{year}{2021}\natexlab{}.
\newblock \showarticletitle{Nerf: Representing scenes as neural radiance fields for view synthesis}.
\newblock \bibinfo{journal}{\emph{Commun. ACM}} \bibinfo{volume}{65}, \bibinfo{number}{1} (\bibinfo{year}{2021}), \bibinfo{pages}{99--106}.
\newblock


\bibitem[Mur-Artal et~al\mbox{.}(2015)]%
        {mur2015orb}
\bibfield{author}{\bibinfo{person}{Raul Mur-Artal}, \bibinfo{person}{Jose Maria~Martinez Montiel}, {and} \bibinfo{person}{Juan~D Tardos}.} \bibinfo{year}{2015}\natexlab{}.
\newblock \showarticletitle{ORB-SLAM: a versatile and accurate monocular SLAM system}.
\newblock \bibinfo{journal}{\emph{IEEE transactions on robotics}} \bibinfo{volume}{31}, \bibinfo{number}{5} (\bibinfo{year}{2015}), \bibinfo{pages}{1147--1163}.
\newblock


\bibitem[Niemeyer et~al\mbox{.}(2022)]%
        {niemeyer2022regnerf}
\bibfield{author}{\bibinfo{person}{Michael Niemeyer}, \bibinfo{person}{Jonathan~T Barron}, \bibinfo{person}{Ben Mildenhall}, \bibinfo{person}{Mehdi~SM Sajjadi}, \bibinfo{person}{Andreas Geiger}, {and} \bibinfo{person}{Noha Radwan}.} \bibinfo{year}{2022}\natexlab{}.
\newblock \showarticletitle{Regnerf: Regularizing neural radiance fields for view synthesis from sparse inputs}. In \bibinfo{booktitle}{\emph{Proceedings of the IEEE/CVF Conference on Computer Vision and Pattern Recognition}}. \bibinfo{pages}{5480--5490}.
\newblock


\bibitem[Park et~al\mbox{.}(2019)]%
        {park2019deepsdf}
\bibfield{author}{\bibinfo{person}{Jeong~Joon Park}, \bibinfo{person}{Peter Florence}, \bibinfo{person}{Julian Straub}, \bibinfo{person}{Richard Newcombe}, {and} \bibinfo{person}{Steven Lovegrove}.} \bibinfo{year}{2019}\natexlab{}.
\newblock \showarticletitle{Deepsdf: Learning continuous signed distance functions for shape representation}. In \bibinfo{booktitle}{\emph{Proceedings of the IEEE/CVF conference on computer vision and pattern recognition}}. \bibinfo{pages}{165--174}.
\newblock


\bibitem[Schonberger and Frahm(2016)]%
        {schonberger2016structure}
\bibfield{author}{\bibinfo{person}{Johannes~L Schonberger} {and} \bibinfo{person}{Jan-Michael Frahm}.} \bibinfo{year}{2016}\natexlab{}.
\newblock \showarticletitle{Structure-from-motion revisited}. In \bibinfo{booktitle}{\emph{Proceedings of the IEEE conference on computer vision and pattern recognition}}. \bibinfo{pages}{4104--4113}.
\newblock


\bibitem[Snavely et~al\mbox{.}(2008)]%
        {snavely2008modeling}
\bibfield{author}{\bibinfo{person}{Noah Snavely}, \bibinfo{person}{Steven~M Seitz}, {and} \bibinfo{person}{Richard Szeliski}.} \bibinfo{year}{2008}\natexlab{}.
\newblock \showarticletitle{Modeling the world from internet photo collections}.
\newblock \bibinfo{journal}{\emph{International journal of computer vision}}  \bibinfo{volume}{80} (\bibinfo{year}{2008}), \bibinfo{pages}{189--210}.
\newblock


\bibitem[Song et~al\mbox{.}(2024)]%
        {song2024darf}
\bibfield{author}{\bibinfo{person}{Jiuhn Song}, \bibinfo{person}{Seonghoon Park}, \bibinfo{person}{Honggyu An}, \bibinfo{person}{Seokju Cho}, \bibinfo{person}{Min-Seop Kwak}, \bibinfo{person}{Sungjin Cho}, {and} \bibinfo{person}{Seungryong Kim}.} \bibinfo{year}{2024}\natexlab{}.
\newblock \showarticletitle{D{\"a}RF: Boosting Radiance Fields from Sparse Input Views with Monocular Depth Adaptation}.
\newblock \bibinfo{journal}{\emph{Advances in Neural Information Processing Systems}}  \bibinfo{volume}{36} (\bibinfo{year}{2024}).
\newblock


\bibitem[Sucar et~al\mbox{.}(2021)]%
        {sucar2021imap}
\bibfield{author}{\bibinfo{person}{Edgar Sucar}, \bibinfo{person}{Shikun Liu}, \bibinfo{person}{Joseph Ortiz}, {and} \bibinfo{person}{Andrew~J Davison}.} \bibinfo{year}{2021}\natexlab{}.
\newblock \showarticletitle{imap: Implicit mapping and positioning in real-time}. In \bibinfo{booktitle}{\emph{Proceedings of the IEEE/CVF International Conference on Computer Vision}}. \bibinfo{pages}{6229--6238}.
\newblock


\bibitem[Sun et~al\mbox{.}(2021)]%
        {sun2021loftr}
\bibfield{author}{\bibinfo{person}{Jiaming Sun}, \bibinfo{person}{Zehong Shen}, \bibinfo{person}{Yuang Wang}, \bibinfo{person}{Hujun Bao}, {and} \bibinfo{person}{Xiaowei Zhou}.} \bibinfo{year}{2021}\natexlab{}.
\newblock \showarticletitle{LoFTR: Detector-free local feature matching with transformers}. In \bibinfo{booktitle}{\emph{Proceedings of the IEEE/CVF conference on computer vision and pattern recognition}}. \bibinfo{pages}{8922--8931}.
\newblock


\bibitem[Teed and Deng(2020)]%
        {teed2020raft}
\bibfield{author}{\bibinfo{person}{Zachary Teed} {and} \bibinfo{person}{Jia Deng}.} \bibinfo{year}{2020}\natexlab{}.
\newblock \showarticletitle{Raft: Recurrent all-pairs field transforms for optical flow}. In \bibinfo{booktitle}{\emph{Computer Vision--ECCV 2020: 16th European Conference, Glasgow, UK, August 23--28, 2020, Proceedings, Part II 16}}. Springer, \bibinfo{pages}{402--419}.
\newblock


\bibitem[Teed and Deng(2021)]%
        {teed2021droid}
\bibfield{author}{\bibinfo{person}{Zachary Teed} {and} \bibinfo{person}{Jia Deng}.} \bibinfo{year}{2021}\natexlab{}.
\newblock \showarticletitle{Droid-slam: Deep visual slam for monocular, stereo, and rgb-d cameras}.
\newblock \bibinfo{journal}{\emph{Advances in neural information processing systems}}  \bibinfo{volume}{34} (\bibinfo{year}{2021}), \bibinfo{pages}{16558--16569}.
\newblock


\bibitem[Truong et~al\mbox{.}(2023)]%
        {truong2023sparf}
\bibfield{author}{\bibinfo{person}{Prune Truong}, \bibinfo{person}{Marie-Julie Rakotosaona}, \bibinfo{person}{Fabian Manhardt}, {and} \bibinfo{person}{Federico Tombari}.} \bibinfo{year}{2023}\natexlab{}.
\newblock \showarticletitle{Sparf: Neural radiance fields from sparse and noisy poses}. In \bibinfo{booktitle}{\emph{Proceedings of the IEEE/CVF Conference on Computer Vision and Pattern Recognition}}. \bibinfo{pages}{4190--4200}.
\newblock


\bibitem[Wang et~al\mbox{.}(2023b)]%
        {wang2023digging}
\bibfield{author}{\bibinfo{person}{Chen Wang}, \bibinfo{person}{Jiadai Sun}, \bibinfo{person}{Lina Liu}, \bibinfo{person}{Chenming Wu}, \bibinfo{person}{Zhelun Shen}, \bibinfo{person}{Dayan Wu}, \bibinfo{person}{Yuchao Dai}, {and} \bibinfo{person}{Liangjun Zhang}.} \bibinfo{year}{2023}\natexlab{b}.
\newblock \showarticletitle{Digging into depth priors for outdoor neural radiance fields}. In \bibinfo{booktitle}{\emph{Proceedings of the 31st ACM International Conference on Multimedia}}. \bibinfo{pages}{1221--1230}.
\newblock


\bibitem[Wang et~al\mbox{.}(2023c)]%
        {wang2023co}
\bibfield{author}{\bibinfo{person}{Hengyi Wang}, \bibinfo{person}{Jingwen Wang}, {and} \bibinfo{person}{Lourdes Agapito}.} \bibinfo{year}{2023}\natexlab{c}.
\newblock \showarticletitle{Co-slam: Joint coordinate and sparse parametric encodings for neural real-time slam}. In \bibinfo{booktitle}{\emph{Proceedings of the IEEE/CVF Conference on Computer Vision and Pattern Recognition}}. \bibinfo{pages}{13293--13302}.
\newblock


\bibitem[Wang et~al\mbox{.}(2023a)]%
        {wang2023f2}
\bibfield{author}{\bibinfo{person}{Peng Wang}, \bibinfo{person}{Yuan Liu}, \bibinfo{person}{Zhaoxi Chen}, \bibinfo{person}{Lingjie Liu}, \bibinfo{person}{Ziwei Liu}, \bibinfo{person}{Taku Komura}, \bibinfo{person}{Christian Theobalt}, {and} \bibinfo{person}{Wenping Wang}.} \bibinfo{year}{2023}\natexlab{a}.
\newblock \showarticletitle{F2-nerf: Fast neural radiance field training with free camera trajectories}. In \bibinfo{booktitle}{\emph{Proceedings of the IEEE/CVF Conference on Computer Vision and Pattern Recognition}}. \bibinfo{pages}{4150--4159}.
\newblock


\bibitem[Wang et~al\mbox{.}(2021)]%
        {wang2021nerf}
\bibfield{author}{\bibinfo{person}{Zirui Wang}, \bibinfo{person}{Shangzhe Wu}, \bibinfo{person}{Weidi Xie}, \bibinfo{person}{Min Chen}, {and} \bibinfo{person}{Victor~Adrian Prisacariu}.} \bibinfo{year}{2021}\natexlab{}.
\newblock \showarticletitle{NeRF--: Neural radiance fields without known camera parameters}.
\newblock \bibinfo{journal}{\emph{arXiv preprint arXiv:2102.07064}} (\bibinfo{year}{2021}).
\newblock


\bibitem[Warburg et~al\mbox{.}(2023)]%
        {warburg2023nerfbusters}
\bibfield{author}{\bibinfo{person}{Frederik Warburg}, \bibinfo{person}{Ethan Weber}, \bibinfo{person}{Matthew Tancik}, \bibinfo{person}{Aleksander Holynski}, {and} \bibinfo{person}{Angjoo Kanazawa}.} \bibinfo{year}{2023}\natexlab{}.
\newblock \bibinfo{title}{Nerfbusters: Removing Ghostly Artifacts from Casually Captured NeRFs}.
\newblock
\newblock
\showeprint[arxiv]{2304.10532}~[cs.CV]


\bibitem[Wei et~al\mbox{.}(2023)]%
        {10378404}
\bibfield{author}{\bibinfo{person}{Tong Wei}, \bibinfo{person}{Yash Patel}, \bibinfo{person}{Alexander Shekhovtsov}, \bibinfo{person}{Jiří Matas}, {and} \bibinfo{person}{Daniel Barath}.} \bibinfo{year}{2023}\natexlab{}.
\newblock \showarticletitle{Generalized Differentiable RANSAC}. In \bibinfo{booktitle}{\emph{2023 IEEE/CVF International Conference on Computer Vision (ICCV)}}. \bibinfo{pages}{17603--17614}.
\newblock
\urldef\tempurl%
\url{https://doi.org/10.1109/ICCV51070.2023.01618}
\showDOI{\tempurl}


\bibitem[Wu(2013)]%
        {wu2013towards}
\bibfield{author}{\bibinfo{person}{Changchang Wu}.} \bibinfo{year}{2013}\natexlab{}.
\newblock \showarticletitle{Towards linear-time incremental structure from motion}. In \bibinfo{booktitle}{\emph{2013 International Conference on 3D Vision-3DV 2013}}. IEEE, \bibinfo{pages}{127--134}.
\newblock


\bibitem[Yan et~al\mbox{.}(2023)]%
        {yan2023cfnerf}
\bibfield{author}{\bibinfo{person}{Qingsong Yan}, \bibinfo{person}{Qiang Wang}, \bibinfo{person}{Kaiyong Zhao}, \bibinfo{person}{Jie Chen}, \bibinfo{person}{Bo Li}, \bibinfo{person}{Xiaowen Chu}, {and} \bibinfo{person}{Fei Deng}.} \bibinfo{year}{2023}\natexlab{}.
\newblock \bibinfo{title}{CF-NeRF: Camera Parameter Free Neural Radiance Fields with Incremental Learning}.
\newblock
\newblock
\showeprint[arxiv]{2312.08760}~[cs.CV]


\bibitem[Zhu et~al\mbox{.}(2022)]%
        {zhu2022nice}
\bibfield{author}{\bibinfo{person}{Zihan Zhu}, \bibinfo{person}{Songyou Peng}, \bibinfo{person}{Viktor Larsson}, \bibinfo{person}{Weiwei Xu}, \bibinfo{person}{Hujun Bao}, \bibinfo{person}{Zhaopeng Cui}, \bibinfo{person}{Martin~R Oswald}, {and} \bibinfo{person}{Marc Pollefeys}.} \bibinfo{year}{2022}\natexlab{}.
\newblock \showarticletitle{Nice-slam: Neural implicit scalable encoding for slam}. In \bibinfo{booktitle}{\emph{Proceedings of the IEEE/CVF Conference on Computer Vision and Pattern Recognition}}. \bibinfo{pages}{12786--12796}.
\newblock


\end{thebibliography}

\par \vfill
\clearpage
\begin{figure*}[!htbp]

     \centering
     \includegraphics[width=0.9\linewidth]{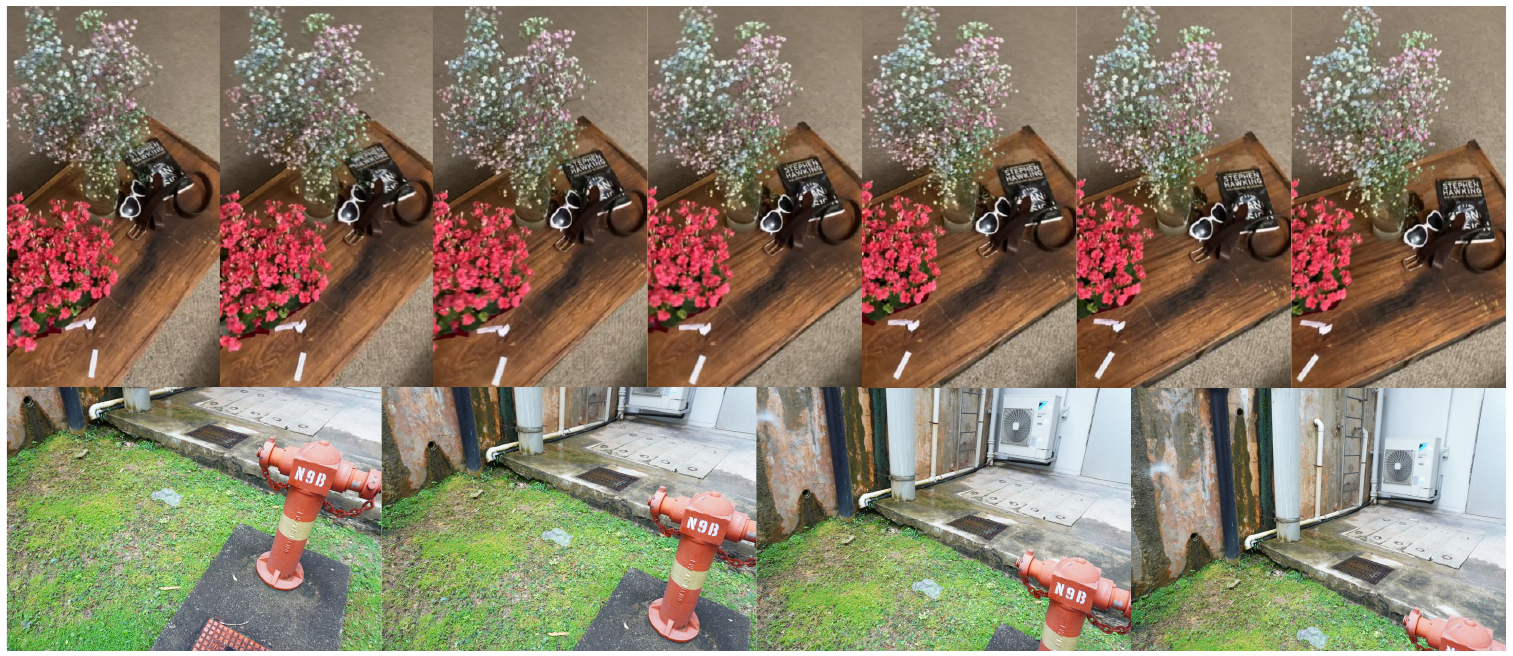}
     \caption{Consecutive frames in the NeRFBuster dataset (top) and the Free Dataset (bottom). The Free Dataset exhibits more pronounced camera motion, posing greater challenges.}
\figlabel{framerate}
\Description{Enjoying the baseball game from the third-base
  seats. Ichiro Suzuki preparing to bat.}
\end{figure*}

\appendix

\section{More Implementation Details}
Our network architecture follows the BARF~\cite{lin2021barf} approach, utilizing a single 8-layer MLP network with a width of 128. All SOTA methods employ their official open-source implementations. For test-optimization, NoPe-NeRF adopts its official implementation, while all other methods undergo 100 iterations of test-optimization after per-image neighbor initialization before evaluation. The Sim(3) alignment approach is also derived from the official open-source version of BARF.


\noindent \textbf{Dataset} We choose two datasets NeRFBuster~\cite{warburg2023nerfbusters} which used in CF-NeRF~\cite{yan2023cfnerf} and Free-Dataset~\cite{wang2023f2} which consists of more challenging scenarios with arbitrary trajectory variations and reduced frame overlap as shown in \figref{framerate}. We utilize the NeRFBuster sequences processed by CF-NeRF. For each scene, CF-NeRF selects approximately 50 images based on their overlap, ordered sequentially. Regarding the Free-Dataset, the \textbf{sky} scene comprises images with indexes from 50 to 100, while all other scenes consist of images with indexes from 0 to 50. All selected sequences present considerable challenges.

\begin{table*}[!htbp]
    \caption{Pose accuracy aligned by approaches of BARF and CF-NeR. $\Delta T$ is the transition error in ground truth scale and  $\Delta R$ is rotation error in degree. Sorted in descending order by $\Delta R$.}
    \tablelabel{align}
    \centering
    \scriptsize
    
    \begin{tabular}{c|c|ccccccccccccc}
    \toprule
         Metrics & Method & aloe & art & car & century & garbage & flowers & picnic & pikachu & pipe & plant & roses & table & mean \\ \midrule
         \multirow{5}{*}{$\Delta R$↓} & CF-NeRF~\cite{yan2023cfnerf} Sim(3) with rotation & 21.918 & 25.702 & 22.653 & 11.245 & 9.061 & 9.915 & 13.489 & 12.046 & 173.343 & 11.056 & 7.002 & 3.837 & 26.772 \\
         & CF-NeRF Sim(3) & 6.703&  76.306&  29.079&  11.013&  74.163&  10.672&  109.868&  13.243&  122.345&  18.664&  3.903 &  3.835 & 39.983\\
         & Ours Sim(3) with rotation & 3.618 & \textbf{0.469} & \textbf{0.545} & \textbf{2.237} & 0.921 & 0.596 & 2.118 & 7.698 & 2.320 & \textbf{5.212} & 1.919 & 1.223 & \textbf{2.406}\\
         & Ours Sim(3) & \textbf{3.163} & 3.151 & 0.701 & 2.343 & \textbf{0.902} & \textbf{0.481} & \textbf{1.938} & \textbf{7.708} & \textbf{2.302} & 6.008 & \textbf{0.570} & \textbf{1.154} & 2.560\\
         \midrule
         \multirow{5}{*}{$\Delta T$↓} & CF-NeRF Sim(3) with rotation & 3.858 & 5.064 & 8.423 & 3.655 & 4.018 & 4.305 & 3.372 & 4.949 & 36.930 & 5.154 & 1.130 & 2.200 & 6.921 \\
         & CF-NeRF Sim(3) & 0.637 & 1.549 & 1.621 & 0.497 & 0.548 & 0.745 & 1.285 &  0.879 &  5.757 &  0.685 &  0.182 & 0.274 & 1.222 \\
         & Ours Sim(3) with rotation & 0.701 & 0.237 & 0.086 & 0.517 & 0.256 & 0.215 & 0.347 &  0.983 &  0.515 &  0.519 &  0.371 & 0.247 & 0.416\\
         & Ours Sim(3) &\textbf{0.168} & \textbf{0.030} & \textbf{0.035} & \textbf{0.134} & \textbf{0.039} & \textbf{0.039} & \textbf{0.106} & \textbf{0.548} & \textbf{0.164} & \textbf{0.225} & \textbf{0.038} & \textbf{0.045} & \textbf{0.131} \\
         \bottomrule
    \end{tabular}
    
\end{table*}

\begin{table*}[!htbp]
    \caption{Comparison to NeRF with COLMAP(GT) pose. }
    \tablelabel{cmp}
    \centering

    \scriptsize
    
    \begin{tabular}{c|ccccc|ccc ccc}
    \toprule
         \multirow{2}{*}{scenes} & \multicolumn{5}{c}{Ours} & \multicolumn{3}{c}{NeRF + Our pose} & \multicolumn{3}{c}{NeRF + COLMAP pose} \\
         \cmidrule{2-6} \cmidrule{7-9} \cmidrule{10-12}
         & $\Delta R$↓ & $\Delta T$↓ & $PSNR$↑ & $SSIM$↑ & $LPIPS$↓ & $PSNR$↑ & $SSIM$↑ & $LPIPS$↓ & $PSNR$↑ & $SSIM$↑ & $LPIPS$↓   \\ \midrule
         pikachu & 7.708 & 0.548 & 22.91 & 0.76 & 0.29 & 23.62 & 0.79 & 0.28 & \textbf{37.06} & \textbf{0.97} & \textbf{0.05} \\
         plant & 6.302 & 0.225 & 22.64 & 0.71 & 0.30 & 20.51 & 0.63 & 0.39 & \textbf{28.27} & \textbf{0.85} & \textbf{0.24} \\
         aloe & 3.163 & 0.168 & 24.36 & 0.61 & 0.35 & \textbf{25.51} & \textbf{0.68} & \textbf{0.26} & 24.04 & 0.58 & 0.40 \\
         art & 3.151 & 0.030 & 26.73 & 0.83 & 0.14 & \textbf{13.77} & \textbf{0.34} & \textbf{0.56} & 12.90 & 0.30 & 0.60 \\
         century & 2.343 & 0.134 & 22.56 & 0.65 & 0.50 & 14.08 & 0.25 & 0.69 & \textbf{14.28} & \textbf{0.28} & \textbf{0.68} \\
         pipe & 2.302 & 0.164 & 23.13 & 0.59 & 0.40 & 21.90 & 0.55 & 0.43 & \textbf{23.05} & \textbf{0.63} & \textbf{0.37} \\
         picnic & 1.938 & 0.106 & 23.04 & 0.52 & 0.49 & 22.52 & 0.51 & 0.45 & \textbf{25.25} & \textbf{0.67} & \textbf{0.31} \\
         table & 1.154 & 0.045 & 32.73 & 0.91 & 0.17 & \textbf{25.64} & \textbf{0.84} & 0.28 & 23.82 & 0.82 & \textbf{0.27} \\
         flowers & 0.902 & 0.039 & 22.69 & 0.63 & 0.43 & \textbf{16.29} & \textbf{0.30} & \textbf{0.66} & 15.38 & 0.27 & 0.67 \\
         car & 0.701 & 0.035 & 27.41 & 0.79 & 0.25 & \textbf{21.52} & \textbf{0.67} & \textbf{0.33} & 18.93 & 0.60 & 0.40 \\
         roses & 0.570 & 0.038 & 29.63 & 0.88 & 0.09 & \textbf{30.09} &	\textbf{0.90} & \textbf{0.09} & 27.77 & 0.84 & 0.16 \\
         garbage & 0.481 & 0.039 & 27.37 & 0.81 & 0.22 & \textbf{18.34} & \textbf{0.59} & \textbf{0.41} & 13.53 & 0.36 & 0.67 \\
         \midrule
         mean & 2.560 & 0.131 & 25.43 & 0.72 & 0.30 & 21.14 & 0.59 & \textbf{0.40} & \textbf{22.02} & \textbf{0.60} & \textbf{0.40} \\

         \bottomrule
    \end{tabular}
    
\end{table*}

\begin{figure*}[!htbp]
     \centering
     \includegraphics[width=\textwidth]{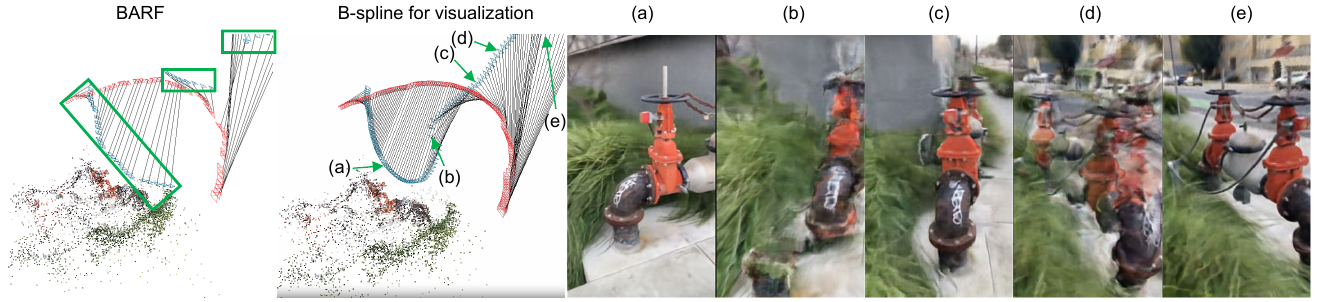}
     \caption{Visualization of segments and interpolated views. (a), (c), and (e) are novel views in segments. (b) and (d) are interpolated novel views}
\figlabel{visoverfit}
\Description{Enjoying the baseball game from the third-base
  seats. Ichiro Suzuki preparing to bat.}
\end{figure*}

\begin{figure*}[!htbp]
     \centering
     \includegraphics[width=\textwidth]{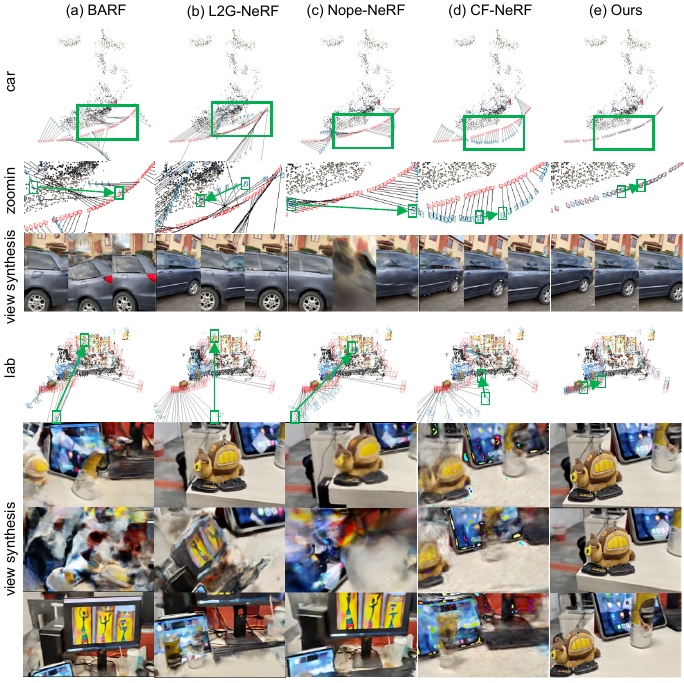}
     \caption{Trajectory comparison. We visualize camera poses of both estimated (blue) and COLMAP (red). Sparse 3D points for the scenes are from COLMAP. While there are abrupt changes in the trajectories of BARF, L2G-NeRF, and Nope-NeRF, the changes are steady along the trajectories of CF-NeRF and ours. The bottom row shows rendered interframes between two frames of abrupt changes denoted by green rectangles.}
\figlabel{overfitting}
\Description{Enjoying the baseball game from the third-base
  seats. Ichiro Suzuki preparing to bat.}
\end{figure*}

\begin{table*}[!ht]
    \caption{Evaluations of the pose accuracy (top 2 rows) and the novel view quality (bottom 3 rows) on Free-Dataset~\cite{wang2023f2}. $\Delta T$ is the transition error in ground truth scale and  $\Delta R$ is rotation error in degree.}
    \tablelabel{free}
    \centering
    
    \setlength{\tabcolsep}{2mm}{
    \scriptsize
    
    \begin{tabular}{c|c|cccccccc}
    \toprule
         Metrics & Method & grass & hydrant & lab & pillar & road & sky & stair & mean \\ \midrule
         \multirow{5}{*}{$\Delta R$↓} & BARF~\cite{lin2021barf} & 124.875 & 74.091 & 124.754 & 16.908 & 64.433 & 22.197 & 0.425 & 61.098 \\
         & L2G-NeRF~\cite{chen2023local} & 114.356 & 170.250 & 56.227 & 131.588 & 109.558 & 27.245 & 162.898 & 110.303 \\
         & Nope-NeRF~\cite{bian2023nope} & 158.408 & 140.245 & 165.086 & 153.613 & 144.478 & 67.749 & 179.836 & 144.202 \\
         & CF-NeRF~\cite{yan2023cfnerf} & 36.875 & 36.129 & 150.882 & 18.282 & 49.790 & 94.082 & 1.260 & 55.329 \\
         & Ours & \textbf{7.785} & \textbf{0.454} & \textbf{6.126} & \textbf{0.124} & \textbf{3.001} & \textbf{2.054} & \textbf{0.089} & \textbf{2.805}\\
         \midrule
         \multirow{5}{*}{$\Delta T$↓} & BARF & 7.273 & 3.890 & 6.675 & 1.475 & 4.587 & 0.583 & \textbf{0.004} & 3.498 \\
         & L2G-NeRF & 7.962 & 7.203 & 4.786 & 7.707 & 7.107 & 0.849 & 10.498 & 6.587 \\
         & Nope-NeRF & 2.920 & 4.904 & 2.062 & 4.044 & 7.201 & 1.156 & 10.564 & 4.693 \\
         & CF-NeRF & 2.850 & 2.018 & 5.998 & 1.382 & 1.808 & 2.518 & 0.121 & 2.385 \\
         & Ours &\textbf{0.304} & \textbf{0.032} & \textbf{0.526} & \textbf{0.008} & \textbf{0.201} & \textbf{0.046} & 0.007 & \textbf{0.161} \\
         \midrule
         \multirow{5}{*}{$PSNR$↑} & BARF & 18.00 & 17.33 & 18.73 & 20.17 & 19.01 & 15.66 & 28.00 & 19.56\\
         & L2G-NeRF & \textbf{18.29} & 17.46 & \textbf{21.18} & 19.93 & 20.49 & 17.90 & 24.41 & 19.95\\
         & Nope-NeRF & 17.02 & 18.33 & 17.55 & 18.99 & 19.08 & 15.39 & 24.35 & 18.67\\
         & CF-NeRF & 18.15 & 17.85 & 16.25 & 20.25 & 18.85 & 15.23 & 21.51 &  18.30\\
         & Ours neighbor & 17.57 & 17.95 & 17.94 & 21.91 & 19.30 & 15.12 & 21.07 & 18.69\\
         & Ours Sim(3) & 16.96 & \textbf{22.54} & 14.88 & \textbf{26.23} & \textbf{24.06} & \textbf{24.37} & \textbf{28.16} & \textbf{22.46}\\
         \midrule
         \multirow{5}{*}{$SSIM$↑} & BARF & 0.40 & 0.33 & 0.58 & 0.52 & 0.47 & 0.49 & 0.83 & 0.52 \\
         & L2G-NeRF & \textbf{0.42} & 0.32 & \textbf{0.67} & 0.53 & 0.52 & 0.55 & 0.74 & 0.54 \\
         & Nope-NeRF & 0.40 & 0.37 & 0.63 & 0.47 & 0.44 & 0.60 & 0.66 &  0.51 \\
         & CF-NeRF & 0.34 & 0.30 & 0.44 & 0.44 & 0.40 & 0.43 & 0.56 & 0.42 \\
         & Ours neighbor& 0.40 & 0.35 & 0.53 & 0.56 & 0.49 & 0.45 & 0.64 & 0.49 \\
         & Ours Sim(3) & 0.36 & \textbf{0.50} & 0.41 & \textbf{0.67} & \textbf{0.61} & \textbf{0.77} & \textbf{0.83} & \textbf{0.59} \\
         \midrule
         \multirow{5}{*}{$LPIPS$↓} & BARF & \textbf{0.51} & 0.60 & 0.38 & 0.50 & 0.51 & 0.48 & \textbf{0.18} & 0.45 \\
         & L2G-NeRF & \textbf{0.51} & 0.61 & \textbf{0.26} & 0.51 & 0.57 & 0.41 & 0.25 & 0.45 \\
         & Nope-NeRF & 0.75 & 0.69 & 0.56 & 0.68 & 0.79 & 0.63 & 0.52 & 0.66 \\
         & CF-NeRF & 0.77 & 0.82 & 0.57 & 0.73 & 0.83 & 0.70 & 0.59 & 0.72 \\
         & Ours neighbor & 0.59 & 0.57 & 0.39 & 0.42 & 0.65 & 0.50 & 0.30 & 0.49 \\
         & Ours Sim(3) & 0.61 & \textbf{0.50} & 0.55 & \textbf{0.37} & \textbf{0.48} & \textbf{0.31} & 0.21 & \textbf{0.43} \\\bottomrule
    \end{tabular}
    }


\end{table*}


\section{Testing methods}

\yq{As mentioned in the main paper, to calculate the metrics for test images, two sequential steps during testing are required: alignment of trajectories for pose quality assessment and test-time optimization for view synthesis quality assessment.}

\noindent \textbf{Alignment} A 3D similarity transformation Sim(3) for the scene and the cameras can be obtained through different methods.
\begin{itemize}
    \item \textbf{Sim(3)} BARF~\cite{lin2021barf} and L2G-NeRF~\cite{chen2023local} align estimated poses to the ground truth through Sim(3) obtained by Procrustes analysis on the camera pose locations.
    \item \textbf{Sim(3) with rotation} CF-NeRF~\cite{yan2023cfnerf} finds that the Procrustes analysis used for Sim(3) is unreliable when all cameras lie in a line or the camera translation contains noise. To overcome the problem, CF-NeRF adds a virtual point $(0, 0, 1)$ in the camera coordinate of each image and uses the camera parameter to transform it to the world coordinate, then uses the camera rotation during the alignment process (termed as rotation). However, we find the approach of CF-NeRF will cause more transition errors.
\end{itemize}

We list the results of pose error on \textbf{buster} both aligned by the approach of \textbf{Sim(3)} and \textbf{Sim(3) with rotation} in \tableref{align}. The results show that the approach of \textbf{Sim(3) with rotation} can reduce rotation errors while causing more transition errors. When the accuracy of poses is high, \textbf{Sim(3) with rotation} takes rare benefits on $\Delta R$ but harms to $\Delta T$. As a result, we employ the \textbf{Sim(3)} approach in the main paper for all methods to align two trajectories and then calculate pose errors.

\noindent \textbf{Test-time optimization} Here we outline previous testing methods with different combinations of initialization and test-time optimization.
\begin{itemize}
    \item \textbf{Sim(3) + opt.} In BARF~\cite{lin2021barf}, the poses are first initialized using Sim(3) alignment with Procrustes analysis on the camera pose locations. Then, an additional test-time optimization is used to further adjust the test poses. This initialization works well when the estimated poses can be aligned precisely to COLMAP poses. However, incorrect pose estimations can affect the Sim(3) alignment.
    \item \textbf{Estimated + no opt.} CF-NeRF~\cite{yan2023cfnerf} recovers all poses without employing a test/train split and then tests every 8th image. However, such an approach leads to results indistinguishable whether the rendered results are due to overfitting or successful reconstruction.
    \item \textbf{Neighbor + opt.} Nope-NeRF~\cite{bian2023nope} initializes the test image pose with the estimated pose of the training frame that is closest to it. Neighbor initialization works well when the framerate is high and the test pose is near the neighbor pose. Facing complex trajectories and reduced overlap it struggles to supply a good initialization as shown in Table 1 in the main paper and \tableref{free}.
\end{itemize}

Due to the substantial alignment errors, all methods except ours struggle to obtain reasonable initial test poses through Sim(3). In our main paper for testing results, we adopt \yq{\textbf{Neighbor + opt.} for all the methods} and also provide results using \yq{\textbf{Sim(3) + opt.}}.

\noindent \textbf{Overfitting} As described in Table 1 and Section 4.2 of the main text and \tableref{free}, although methods like BARF converge to significant pose errors and poor structural quality, they still achieve comparable novel view synthesis metrics to our method. We attribute this to the network converging to local optima. The left part of \figref{visoverfit} illustrates the poses estimated by BARF, where the three green boxes indicate three pose segments after fitting. To render a video, we fit B-spline functions to the estimated poses to get a smooth camera trajectory. The right part visualizes novel views synthesized for poses within the segments (a,c,e) and between segments (b, d) on the B-spline trajectory. The novel views synthesized for poses in the three segments (a, c, e) appear normal. However, the visualization results for the interpolated poses (b,d) between segments are unreasonable. It seems that each segment fits a sub-scene and the images for the interpolated poses between segments stitch different scenes together.
\figref{overfitting} shows more results of this issue. These methods do not recover correct poses and scene geometry. During test-time optimization, poses for testing images are initialized with the estimated poses of neighboring training images. With the twisted scene and fragmented pose trajectories after training, the test-time optimization results in the pose of a test frame converging to a "pseudo " pose close to the estimated pose of the closest neighboring training frame. Then the network for the scene further overfits to the "pseudo" ground truth test pose and image ($P, I$ for example) pairs and renders an image  $I'$ using $P$ to calculate visual metrics with $I$, leading to high view synthesis metrics. Notice that during the process, $P$ diverges far from its true pose to the direction minimizing $I'$ and $I$ when the scene geometry and camera poses exhibit large errors.

\section{Comparison to NeRF with COLMAP pose}
We additionally compare the novel view synthesis quality of the NeRF model trained with our estimated pose and COLMAP pose (we use it as GT) to demonstrate the pose accuracy estimated.
On average, \textbf{NeRF + Our pose} achieves novel view quality close to that of \textbf{NeRF + COLMAP pose}. In some scenes, our poses have large estimation error, like \textbf{Pikachu} and \textbf{plant}. Both pose error during training and test pose misalignment lead to worse view quality of \textbf{NeRF + Our pose} in these scenes.
In scenes with small pose error, \textbf{NeRF + Our pose} gains similar view quality with \textbf{NeRF + COLMAP pose}, even better in many scenes.
In many scenes, our methods achieve better view quality than \textbf{NeRF + Our pose} and \textbf{NeRF + COLMAP pose}. We attribute it to coarse to fine positional encoding, reprojection loss, and joint optimization of poses and scenes.


\section{More Results}
Detailed results of Free-Dataset are shown in \tableref{free}. We provide more qualitative comparisons with state-of-the-art works. \figref{busternv1}, \figref{busternv2}, \figref{busternv3} and \figref{freenv} illustrates a comparison of novel view synthesis quality. \figref{busterdp1}, \figref{busterdp2}, \figref{busterdp3} and \figref{freedp} demonstrates a comparison of depth map rendering quality. \figref{bustertraj1}, \figref{bustertraj2} and \figref{freetraj} present a comparison of the reconstructed trajectories obtained by various methods, where the red boxes represent the COLMAP poses, and the blue boxes depict the estimated poses. The point cloud, processed from the COLMAP output, serves as a reference for relative positioning.

\begin{figure*}[!htbp]
     \centering
     \includegraphics[width=\textwidth]{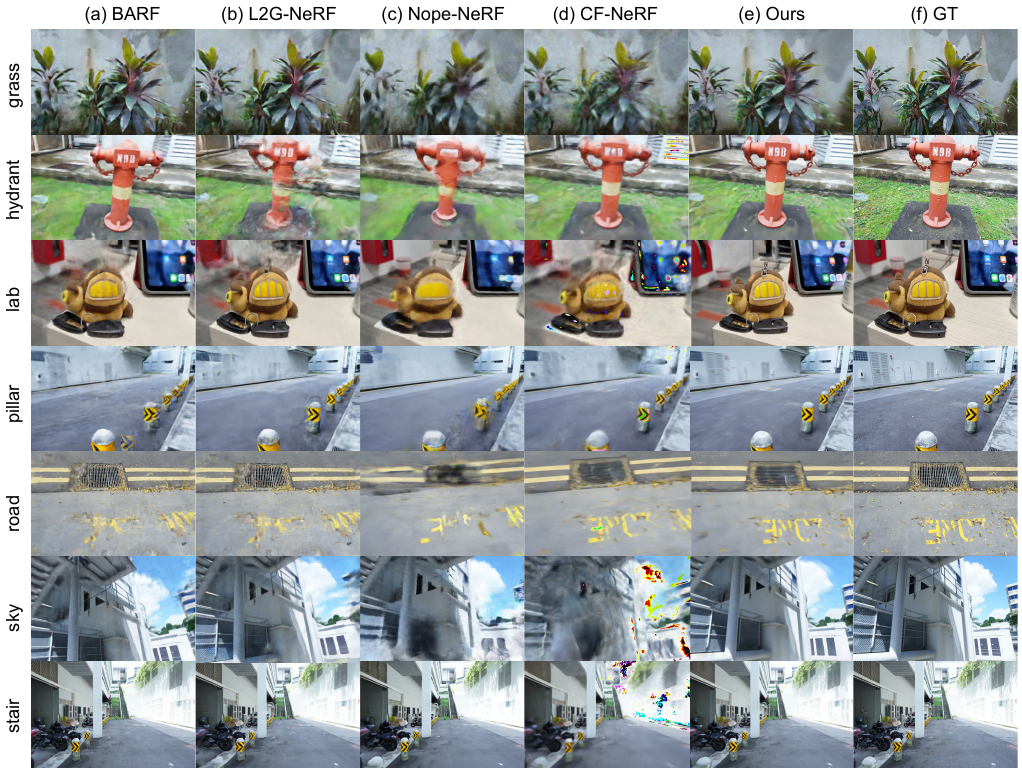}
     \caption{Comparison of novel views on Free-Dataset~\cite{wang2023f2}.}
\figlabel{freenv}
\Description{Enjoying the baseball game from the third-base
  seats. Ichiro Suzuki preparing to bat.}
\end{figure*}

\begin{figure*}[!htbp]
     \centering
     \includegraphics[width=\textwidth]{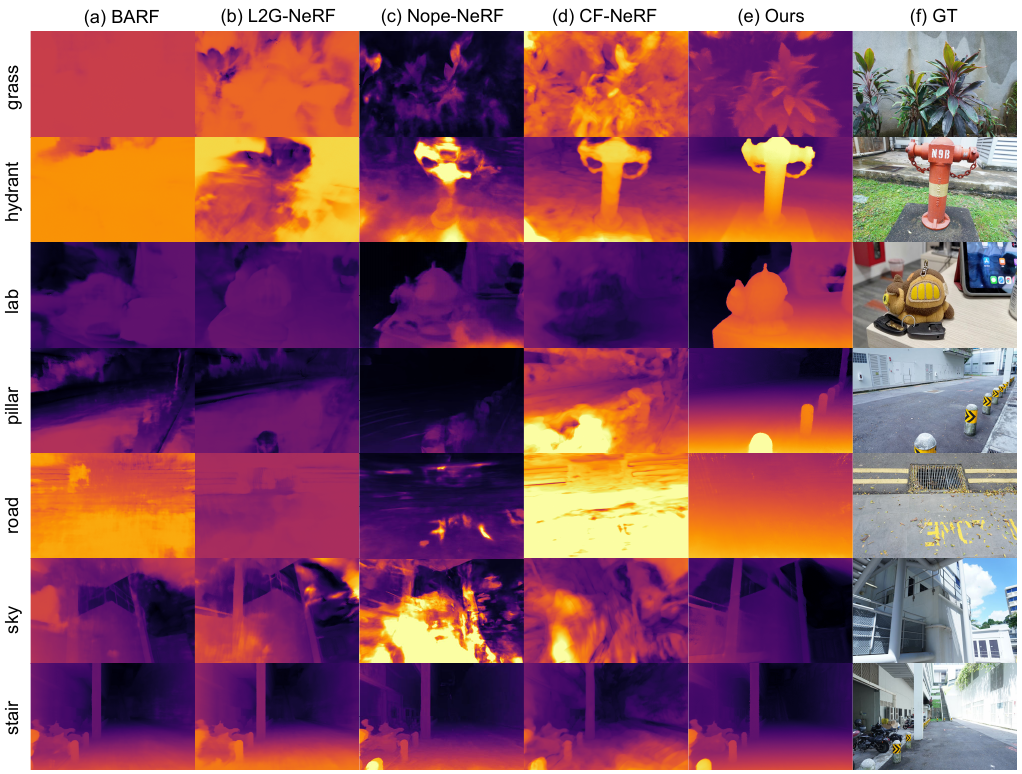}
     \caption{Comparison of rendered depths on Free-Dataset~\cite{wang2023f2}.}
\figlabel{freedp}
\Description{Enjoying the baseball game from the third-base
  seats. Ichiro Suzuki preparing to bat.}
\end{figure*}

\begin{figure*}[!htbp]
     \centering
     \includegraphics[width=0.95\textwidth]{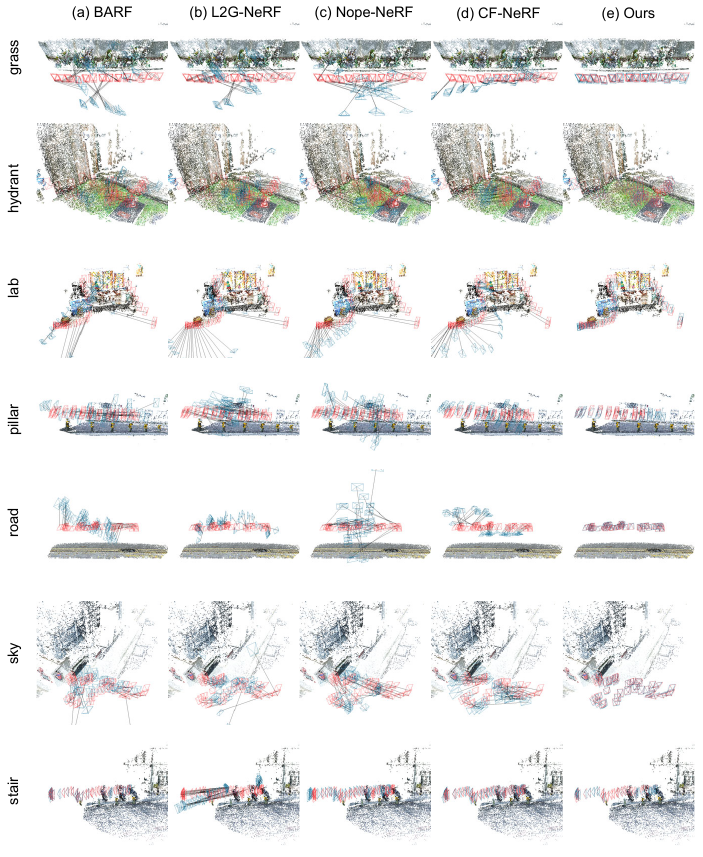}
     \caption{Trajectory comparison on Free-Dataset~\cite{wang2023f2}. We visualize camera poses of both estimated (blue) and COLMAP (red). Sparse 3D points for the scenes are from COLMAP.}
\figlabel{freetraj}
\Description{Enjoying the baseball game from the third-base
  seats. Ichiro Suzuki preparing to bat.}
\end{figure*}

\begin{figure*}[!htbp]
    
     \centering
     \includegraphics[width=0.95\linewidth]{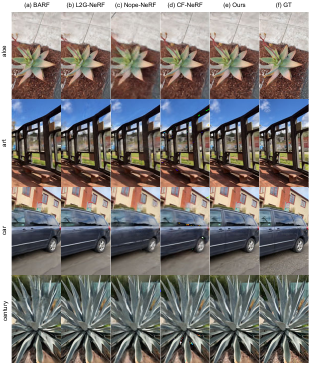}
     \caption{Comparison of novel views on NeRFBuster~\cite{warburg2023nerfbusters}. Part one.}
\figlabel{busternv1}
\Description{Enjoying the baseball game from the third-base
  seats. Ichiro Suzuki preparing to bat.}
\end{figure*}

\begin{figure*}[!htbp]
    
     \centering
     \includegraphics[width=0.95\linewidth]{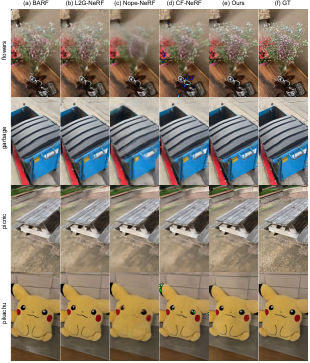}
     \caption{Comparison of novel views on NeRFBuster~\cite{warburg2023nerfbusters}. Part two.}
\figlabel{busternv2}
\Description{Enjoying the baseball game from the third-base
  seats. Ichiro Suzuki preparing to bat.}
\end{figure*}

\begin{figure*}[!htbp]
    
     \centering
     \includegraphics[width=0.95\linewidth]{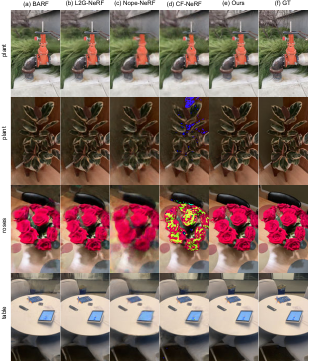}
     \caption{Comparison of novel views on NeRFBuster~\cite{warburg2023nerfbusters}. Part three.}
\figlabel{busternv3}
\Description{Enjoying the baseball game from the third-base
  seats. Ichiro Suzuki preparing to bat.}
\end{figure*}

\begin{figure*}[!htbp]
     \centering
     \includegraphics[width=0.95\linewidth]{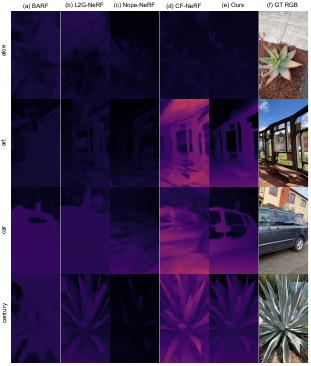}
     \caption{Comparison of rendered depths on NeRFBuster~\cite{warburg2023nerfbusters}. Part one.}
\figlabel{busterdp1}
\Description{Enjoying the baseball game from the third-base
  seats. Ichiro Suzuki preparing to bat.}
\end{figure*}

\begin{figure*}[!htbp]
     \centering
     \includegraphics[width=0.95\textwidth]{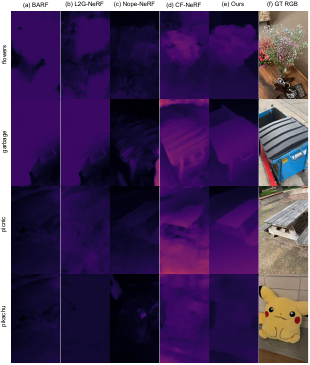}
     \caption{Comparison of rendered depths on NeRFBuster~\cite{warburg2023nerfbusters}. Rendered depths. Part two.}
\figlabel{busterdp2}
\Description{Enjoying the baseball game from the third-base
  seats. Ichiro Suzuki preparing to bat.}
\end{figure*}

\begin{figure*}[!htbp]
     \centering
     \includegraphics[width=0.95\textwidth]{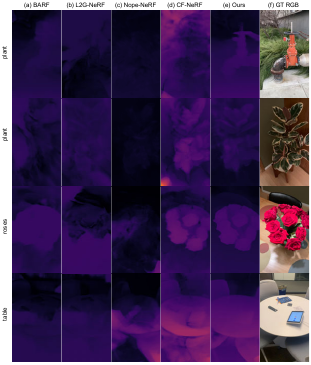}
     \caption{Comparison of rendered depths on NeRFBuster~\cite{warburg2023nerfbusters}. Rendered depths. Part three.}
\figlabel{busterdp3}
\Description{Enjoying the baseball game from the third-base
  seats. Ichiro Suzuki preparing to bat.}
\end{figure*}

\begin{figure*}[!htbp]
     \centering
     \includegraphics[width=0.95\textwidth]{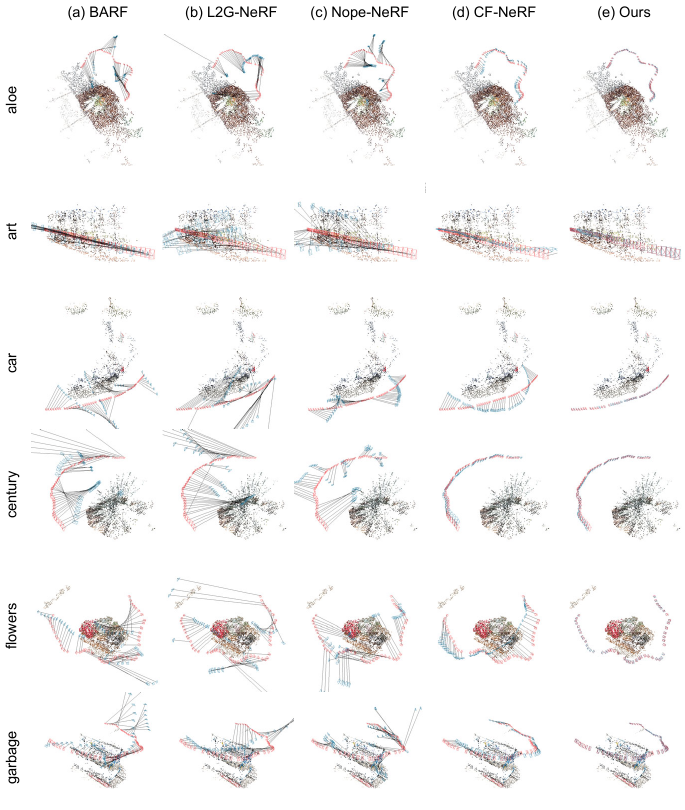}
     \caption{Trajectory comparison on NeRFBuster~\cite{warburg2023nerfbusters}. We visualize camera poses of both estimated (blue) and COLMAP (red). Sparse 3D points for the scenes are from COLMAP. Part one.}
\figlabel{bustertraj1}
\Description{Enjoying the baseball game from the third-base
  seats. Ichiro Suzuki preparing to bat.}
\end{figure*}

\begin{figure*}[!htbp]
     \centering
     \includegraphics[width=0.95\textwidth]{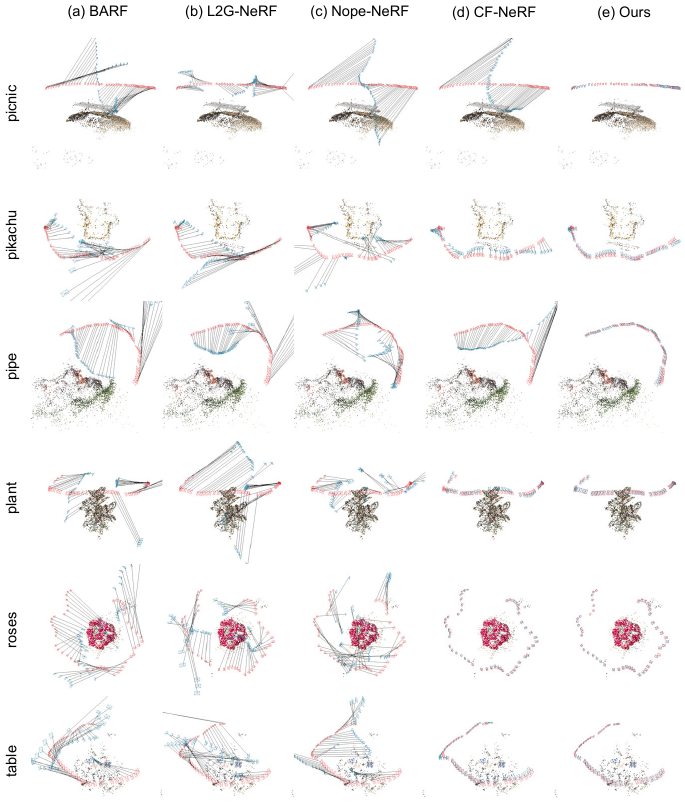}
     \caption{Trajectory comparison on NeRFBuster~\cite{warburg2023nerfbusters}. We visualize camera poses of both estimated (blue) and COLMAP (red). Sparse 3D points for the scenes are from COLMAP. Part two.}
\figlabel{bustertraj2}
\Description{Enjoying the baseball game from the third-base
  seats. Ichiro Suzuki preparing to bat.}
\end{figure*}

\par\vfill\par

\clearpage  



\end{document}